\documentclass[journal]{IEEEtran}

\usepackage{booktabs}
\usepackage{multirow}
\usepackage{graphicx}
\usepackage{amssymb} 
\usepackage{makecell} 
\usepackage{amsmath}
\usepackage{url}
\usepackage{orcidlink}
\usepackage{float}
\usepackage{cite}
\usepackage{placeins}
\usepackage{placeins}
\usepackage{subcaption}
\usepackage{xcolor}
\definecolor{secondbest}{rgb}{0,0.5,0}
\definecolor{best}{HTML}{0072B2}

\ifCLASSINFOpdf
\else
\fi

\begin{document}
\title{LiM-YOLO: Less is More with Pyramid Level Shift for Ship Detection in Optical Remote Sensing}
%
%
%

\author{Seon-Hoon~Kim\orcidlink{0009-0008-9261-6828},
        Yerin~Kim,
        Hyeji~Sim,
        Youeyun~Jung,
        Okchul~Jung,
        and~Daewon~Chung\orcidlink{0000-0003-2638-050X},%
\thanks{Corresponding author: Daewon Chung.}%
\thanks{Seon-Hoon~Kim is with the University of Science and Technology (UST), Daejeon 34113, Republic of Korea (e-mail: egshkim@gmail.com).}%
\thanks{Yerin~Kim, Hyeji~Sim, Youeyun~Jung, Okchul~Jung, and Daewon~Chung are with the Korea Aerospace Research Institute (KARI), Daejeon 34133, Republic of Korea (e-mail: yerin@kari.re.kr; havewisdom@kari.re.kr; yejung@kari.re.kr; ocjung@kari.re.kr; dwchung@kari.re.kr).}%
}


\maketitle
%
%

\markboth{Journal of \LaTeX\ Class Files,~Vol.~13, No.~9, September~2014}%
{Shell \MakeLowercase{\textit{et al.}}: Bare Demo of IEEEtran.cls for Journals}
%



\begin{abstract}
General-purpose object detectors face fundamental structural 
limitations when applied to ship detection in satellite 
imagery, where the ship scale distribution is concentrated 
at small sizes and high aspect ratios. In conventional You Only Look Once architectures, the deepest feature
pyramid level (stride 32) compresses narrow vessels into sub-pixel
representations, causing severe spatial feature dilution and compromising
accurate ship boundary regression. We propose Less is More YOLO, a streamlined
detector built upon the extra-large variant of YOLOv9, to address these
domain-specific structural conflicts. From a statistical analysis of ship scale distributions 
across four major benchmarks (SODA-A, DOTA-v1.5, 
FAIR1M-v2.0, and ShipRSImageNet), we introduce a 
Pyramid Level Shift Strategy that shifts the detection 
head from strides 8, 16, and 32 to strides 4, 8, and 16. This shift satisfies a spatial 
representability condition derived from the Nyquist--Shannon 
principle for the narrowest targets, while eliminating the 
computational redundancy of the 
deepest pyramid level. To further stabilize training on
high-resolution satellite inputs, we incorporate a group-normalized composite-backbone projection module, mitigating gradient instability in memory-constrained micro-batch regimes. Validated on these four datasets, our detector attains an
mAP\textsubscript{50:95} of 0.600 with only 21.16 million parameters, a 64.1\%
reduction from the extra-large YOLOv9 baseline (58.99 million). Despite this
compact size, our model surpasses state-of-the-art detectors up to three times larger, validating that a well-targeted pyramid level shift achieves
a ``Less is More'' balance between accuracy and efficiency. The code is
available at \url{https://github.com/egshkim/LiM-YOLO}.
\end{abstract}


%
\IEEEpeerreviewmaketitle

\section{Introduction}

With the rapid growth of global maritime traffic, automated ship 
detection using high-resolution remote sensing imagery has become 
increasingly important for marine safety, traffic management, and 
maritime law enforcement~\cite{bakirci2025advanced, magalhaes2025vessel, sankhe2024vessel, zhao2024ship}. 
Among various deep learning frameworks, 
the You Only Look Once (YOLO) family~\cite{YOLOv1} 
has gained widespread adoption for this task owing to its favorable 
balance between inference speed and detection accuracy. However, the 
YOLO architecture was originally designed for natural images (e.g., 
MS COCO~\cite{lin2014microsoft}) and carries inherent structural limitations when applied to 
ship detection in satellite imagery.

At the heart of the problem lies the grid-based detection mechanism 
of YOLO, whose prediction granularity is governed by the spatial 
resolution of the feature map~\cite{ObjectDetectionReview}. In the 
Feature Pyramid Network (FPN)~\cite{FPN} convention adopted by the 
YOLO series since YOLOv3~\cite{YOLOv3}, P$n$ denotes the feature 
map at the $n$-th pyramid level with a downsampling stride of $2^n$ 
relative to the input image. From YOLOv3 through the latest 
YOLOv12~\cite{YOLOv12}, almost all variants inherit a three-level 
pyramid comprising P3, P4, and P5, with strides of 
$2^3\!=\!8$, $2^4\!=\!16$, and $2^5\!=\!32$, respectively. While 
this configuration performs well on common objects in natural images, 
it is ill-suited for ships. In the overhead geometry of 
satellite imagery, ships typically appear as narrow, elongated 
structures spanning only a few pixels along their minor axis. 
At the P5 stride of 32, such targets are compressed below the 
resolution of a single grid cell, leading to severe spatial feature 
dilution in which the morphological cues of the object are submerged 
in background content.

Despite this structural mismatch, existing research in ship detection 
has largely focused on improving feature extraction or fusion modules 
within the fixed P3--P5 framework. While some recent works have 
attempted to incorporate higher-resolution pyramid levels such as P2, 
they adopted what we term an ``expansion-only'' strategy, appending 
additional levels to the existing P3--P5 configuration without 
removing any redundant deep layers. Retaining the P5 level not only 
incurs unnecessary computational cost but also introduces an excessively large Effective Receptive Field (ERF) that encodes more background context 
than object-specific information.

In this paper, we challenge the prevailing assumption that deeper 
feature hierarchies necessarily improve detection performance, 
particularly for the narrow, small-scale targets common in maritime 
surveillance. We propose the Less is More YOLO (LiM-YOLO), a streamlined 
architecture developed through a rigorous, data-driven analysis of 
ship scale distributions across multiple datasets. The central 
element of our approach is a Pyramid Level Shift Strategy, which 
reconfigures the detection head from the conventional P3--P5 to a 
P2--P4 structure. By integrating the high-resolution P2 level, we 
ensure that the vast majority of ships occupy at least one full grid 
cell in the feature map, thereby preserving the spatial information 
required for accurate boundary regression. Simultaneously, by pruning 
the redundant P5 backbone and head, we eliminate a major source of 
background content and computational waste, achieving an architecture 
that is both lighter and more accurate.

Although the original YOLOv9 architecture does not employ 
normalization within its Programmable Gradient Information (PGI) 
framework, we find that normalization is essential for stabilizing 
training when processing complex remote sensing data. Training a 
large-scale model on high-resolution satellite imagery, however, 
necessitates micro-batch training due to GPU memory constraints. 
Under such conditions, standard Batch Normalization (BN) suffers 
from unreliable statistical estimates, leading to degraded 
performance. To address this, we introduce GN-CBLinear, which leverages Group Normalization (GN)~\cite{GN}, a normalization scheme independent of batch size, 
to ensure stable gradient flow and convergence even in 
memory-constrained settings.

We validate the proposed method through extensive experiments on 
four diverse datasets: SODA-A~\cite{SODA-A}, DOTA-v1.5~\cite{DOTA}, 
FAIR1M-v2.0~\cite{FAIR1M}, and ShipRSImageNet~\cite{ShipRSImageNet}. 
LiM-YOLO outperforms state-of-the-art models, including YOLOv8~\cite{YOLOv8}, YOLOv10~\cite{YOLOv10}, YOLO11~\cite{yolo11_ultralytics}, YOLOv12~\cite{YOLOv12}, and 
RT-DETR~\cite{RT-DETR}, delivering higher detection accuracy with 
significantly fewer parameters. Qualitative analysis further reveals 
that the enhanced spatial resolving power of LiM-YOLO markedly 
improves the detectability of small and densely packed vessels.

The contributions of this work are summarized as follows.

\begin{enumerate}
\item We conduct a comprehensive statistical analysis of ship 
morphometry across four major remote sensing datasets, 
quantitatively identifying the spatial feature dilution and 
receptive field redundancy induced by the conventional P5 layer 
(stride $2^5=32$).

\item We propose LiM-YOLO, a novel architecture that shifts the 
feature pyramid from P3--P5 to P2--P4, effectively resolving the 
scale mismatch between the detector and ships. This 
``Less is More'' design achieves a favorable balance between 
detection accuracy and computational efficiency.

\item We introduce GN-CBLinear, a 
batch-size-independent normalization of the composite-backbone projection that stabilizes 
training of deep networks on high-resolution satellite imagery, 
overcoming the limitations of BN in micro-batch 
regimes.

\item Through extensive ablation studies on four diverse 
benchmarks and head-to-head comparisons against extra-large 
variants of contemporary YOLO and transformer-based detector 
families, we empirically validate the effectiveness of the 
proposed pyramid level shift. LiM-YOLO achieves superior 
accuracy with substantially fewer parameters than these 
larger competing detectors, demonstrating effectiveness of well-targeted pyramid design.
\end{enumerate}

\section{Related Work}

\subsection{Evolution of YOLO Architecture}

The YOLO family has undergone continuous development aimed at improving the balance between inference speed 
and detection accuracy. Since YOLOv3~\cite{YOLOv3}, the three-level P3--P5 head has 
become the de facto standard across nearly all YOLO variants. 
Within this configuration, P3 retains fine-grained spatial 
details suitable for small objects, whereas P5 captures 
high-level semantic information for larger targets.

Subsequent YOLO variants have focused their architectural 
improvements almost exclusively on the backbone and neck rather 
than the head. YOLOv4~\cite{YOLOv4} introduced 
CSPDarknet53, applying Cross-Stage Partial (CSP) connections to improve 
computational efficiency, and YOLOv5~\cite{YOLOv5} refined this 
idea through the C3 module. YOLOv8~\cite{YOLOv8} replaced C3 with 
the C2f module to facilitate richer feature reuse and adopted an 
anchor-free detection paradigm. Other notable developments include 
the hardware-aware design of YOLOv6~\cite{YOLOv6}, the decoupled 
head in YOLOX~\cite{YOLOX}, and the E-ELAN aggregation structure 
in YOLOv7~\cite{YOLOv7}. More recently, 
YOLOv10~\cite{YOLOv10} proposed an NMS-free training strategy, 
YOLO11~\cite{yolo11_ultralytics} incorporated improved C3k2 
blocks, and YOLOv12~\cite{YOLOv12} explored attention-centric 
architectures. In parallel, Zhao et al.~\cite{RT-DETR} introduced 
RT-DETR, a real-time end-to-end transformer detector that 
eliminates the latency associated with Non-Maximum Suppression 
through a hybrid encoder and uncertainty-minimal query selection.

Despite these advances, the vast majority of current models, 
including the latest versions, uncritically inherit the P3, P4, 
and P5 head 
configuration established in YOLOv3. While this setup performs 
adequately on general-purpose benchmarks such as MS COCO, it 
presents a structural limitation for remote sensing ship detection, 
where targets are often exceedingly small and the coarse resolution 
of P5 ($1/32$ of the input) is insufficient for meaningful feature 
representation.

This structural mismatch motivates our investigation of an 
alternative head configuration, for which we select a baseline 
that already addresses related information-preservation challenges 
in deep networks. Among recent YOLO variants, YOLOv9~\cite{YOLOv9} 
is distinguished by introducing PGI together with the 
Generalized Efficient Layer Aggregation Network (GELAN), 
mechanisms designed to mitigate information bottlenecks in deep 
networks and that we leverage and extend in our normalization 
design (Section~\ref{sec:proposed_method}). We therefore adopt its 
extra-large configuration, YOLOv9-E, as our baseline. Beyond this 
architectural alignment, YOLOv9-E remains competitive in absolute 
detection accuracy among contemporary high-capacity detectors. On 
the MS COCO 2017 validation set, it achieves 55.6\% 
mAP~\cite{YOLOv9}, surpassing YOLOv10-X 
(54.4\%)~\cite{YOLOv10}, YOLO11-X (54.6\%)~\cite{yolo11_ultralytics}, 
YOLOv12-X (55.2\%)~\cite{YOLOv12}, and the transformer-based 
RT-DETR-R101 (54.3\%)~\cite{RT-DETR}. This combination of an information-preserving design (GELAN with the composite backbone) and strong general-purpose 
accuracy makes YOLOv9-E a suitable platform on which to explore an 
optimal head design for remote sensing ship detection.

\subsection{YOLO for Ship Detection without Head-Level Modification}
\label{sec:ship_no_head}

A substantial body of work~\cite{LMO-YOLO, YOLOv7-ship, CM-YOLO, ShadowFPN-YOLO} 
has sought to improve ship detection by strengthening the backbone 
and neck while preserving the conventional P3, P4, and P5 detection 
scales. These efforts primarily aim to reduce false positives caused 
by complex maritime backgrounds or to mitigate the erosion of small 
vessel features during downsampling, typically by introducing modules 
that enhance contextual information or suppress background content.

Xu et al.~\cite{LMO-YOLO} proposed LMO-YOLO, which identified the 
sparsity of ship features in low-resolution satellite imagery as a 
key bottleneck. They incorporated a Multi-scale Dilated Convolution 
(MDC) module into the YOLOv4 backbone, expanding the receptive field 
to capture richer contextual relationships between objects and their 
surroundings. Jiang et al.~\cite{YOLOv7-ship} pursued a lightweight 
alternative with YOLOv7-Ship, optimizing YOLOv7-Tiny by embedding a 
Coordinate Attention Mechanism (CA-M) in the backbone to suppress 
background interference and enhancing small-vessel feature 
preservation in the neck through OD-ELAN with Omnidimensional 
Dynamic Convolution (ODConv)~\cite{ODConv} and CARAFE 
upsampling~\cite{CARAFE}.

More recently, methods targeting dense ship detection in complex 
coastal environments have been proposed. CM-YOLO~\cite{CM-YOLO}, 
built upon YOLOX, employs a Dual Path Context Enhancement (DCE) 
module in the neck to extract global context and a Multi-Context 
Boosted (MCB) head to improve scale awareness. Although the internal 
structure of the detection head was modified, the model retains the 
P3, P4, and P5 pyramid levels. Similarly, Yang 
et al.~\cite{ShadowFPN-YOLO} proposed ShadowFPN-YOLO based on 
YOLOv10, integrating GELAN with Reparameterized CSP modules in the 
backbone and applying ShadowFPN in the neck, which randomly masks 
background regions of feature maps to force the network to focus on 
intrinsic ship features. Despite these advances, these studies remain 
confined to optimizing feature extraction and fusion within the fixed 
constraints of the conventional detection scales.

\subsection{YOLO for Ship Detection with Head-Level Modification}
\label{sec:ship_with_head}

Beyond module-level enhancements in the backbone and neck, recent 
work~\cite{YOLO-RSA, YOLO-Ssboat} has sought to improve performance 
by expanding the Feature Pyramid Levels in the detection head. These 
strategies generally fall into two directions. Some append deeper 
levels (e.g., P6) to broaden the receptive field for large targets, 
while others add shallower levels (e.g., P2) to preserve finer 
spatial details of small objects.

Fang et al.~\cite{YOLO-RSA} addressed the multi-scale nature of 
optical remote sensing imagery by proposing YOLO-RSA, which appends 
a P6 detection head to the standard P3--P5 configuration. The 
inclusion of P6 aims to capture a broader receptive field, 
incorporating contextual information from the entire image or 
accommodating extremely large vessels.

In the opposite direction, Zeng et al.~\cite{YOLO-Ssboat} proposed 
YOLO-Ssboat based on YOLOv8, specifically targeting super-small 
vessel detection in drone and satellite imagery. They employed C2f 
modules based on Deformable Convolution 
(DCNv3)~\cite{DCNv3} in the backbone to handle shape variations and 
introduced a Multi-Scale Weighted Pyramid Network (MSWPN) in the 
neck. Their most significant structural change, however, lies in the 
detection head, where they adopted a 4-scale configuration that 
includes the high-resolution P2 level (stride $2^2\!=\!4$) alongside 
P3, P4, and P5.

\subsection{Limitations of Existing Approaches}

Despite the advances reviewed above, structural limitations persist 
in both lines of work. The methods discussed in 
Section~\ref{sec:ship_no_head} perform detection starting from the 
P3 level (stride $2^3\!=\!8$). Consequently, they inherently 
struggle with tiny vessels whose spatial features have already been 
diluted during backbone downsampling. Despite advances in feature 
fusion, spatial information lost during downsampling cannot be 
perfectly recovered.

The head-expansion approaches described in 
Section~\ref{sec:ship_with_head} attempt to overcome this limitation 
but exhibit a common flaw. They expand the head architecture 
heuristically, without a rigorous statistical analysis of the object 
scale distribution within the target domain. Specifically, 
YOLO-RSA~\cite{YOLO-RSA}, which extends the architecture to include 
P6, was validated on the HRSC2016 dataset~\cite{HRSC2016}, a 
benchmark dominated by large warships. Whether the P6 level is 
genuinely beneficial in general maritime environments, which are 
characterized by diverse resolutions and predominantly smaller 
vessels, remains unverified. Similarly, 
YOLO-Ssboat~\cite{YOLO-Ssboat} incorporated P2 to target small 
vessels, yet retained the P5 level (stride $2^5\!=\!32$), which 
contributes little to detecting super-small targets and adds 
computational overhead. In both cases, the existing pyramid levels 
are kept intact, an ``expansion-only'' strategy that fails to align 
the architecture with the actual data distribution.

These observations motivate our approach of simultaneously 
introducing a high-resolution P2 level and removing the redundant 
P5 level, rather than simply expanding the number of pyramid levels. 
The details of this strategy are presented in 
Section~\ref{sec:proposed_method}.

\begin{figure*}[!t]
\centering
\includegraphics[width=\textwidth]{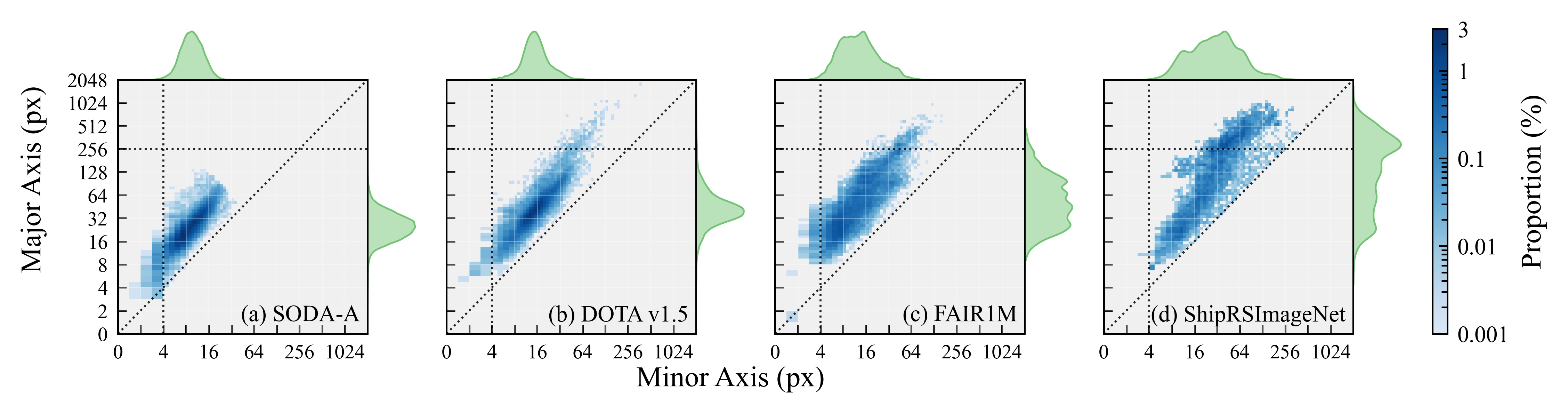}
\caption{Joint distributions of ship minor and major axes 
(in pixels) across four benchmark datasets: 
(a)~SODA-A ($N\!=\!65{,}659$), (b)~DOTA~v1.5 ($N\!=\!43{,}738$), 
(c)~FAIR1M ($N\!=\!58{,}982$), and 
(d)~ShipRSImageNet ($N\!=\!13{,}065$). $N$ denotes the number of 
ship instances in the original datasets used for 
scale-distribution analysis. Patch-level counts after preprocessing 
are listed in Table~\ref{tab:dataset_composition}. Each cell 
represents the per-dataset proportion (\%) of instances whose OBB 
dimensions fall within the corresponding bin, with color intensity 
on a logarithmic scale. Marginal kernel density estimate (KDE) curves along the top and right 
edges show the univariate distributions of minor and major axes, 
respectively. Both axes use a hybrid linear--log$_2$ transform 
(linear below 2\,px, log$_2$ above). Black dotted references: 
vertical at minor axis $=4$\,px (2.5th percentile of minor axis), 
horizontal at major axis $=256$\,px (97.5th percentile of major 
axis), and diagonal at $x\!=\!y$ (isotropic bounding boxes).}
\label{fig:ship_size_distribution}
\end{figure*}

\section{Dataset Analysis}
\label{sec:dataset_analysis}

To comprehensively evaluate the proposed model, we 
selected four benchmark datasets: 
SODA-A~\cite{SODA-A}, DOTA-v1.5~\cite{DOTA}, 
FAIR1M-v2.0~\cite{FAIR1M}, and 
ShipRSImageNet~\cite{ShipRSImageNet}. These datasets 
were chosen to represent a high degree of heterogeneity 
in sensor platforms, Ground Sampling Distance (GSD), 
object scale distributions, and scene contexts. All 
datasets provide Oriented Bounding Box (OBB) annotations 
for ship objects.

\subsection{SODA-A}
SODA-A~\cite{SODA-A} is a large-scale dataset tailored 
for small object detection in aerial imagery, constructed 
primarily from high-resolution Google Earth scenes (GSD 
0.5--0.8\,m, average dimensions exceeding 
$4700 \times 2700$~pixels). Its distinguishing feature 
is the extremely high density of tiny instances, with an 
average object size of approximately 14.75~pixels. We 
extracted only ship instances for this study, yielding 
1,030 training and 323 validation images containing 
37,971 and 21,908 OBB-annotated ships, respectively.

\subsection{DOTA-v1.5}
DOTA-v1.5~\cite{DOTA} is an advanced version of DOTA 
sourced from Google Earth, Gaofen-2 (GF-2), and Jilin-1 
(JL-1), designed to challenge detectors with complex 
aerial scenes containing many small instances (less 
than 10~pixels). Images vary in size from 
$800 \times 800$ to $4000 \times 4000$~pixels. Filtering 
for ship-bearing scenes yielded 2,657 training images (56,313 OBB) and 572 validation images (11,474 OBB).

\subsection{FAIR1M-v2.0}
FAIR1M-v2.0~\cite{FAIR1M} is a large-scale benchmark 
for fine-grained object recognition, constructed 
primarily from the GF satellite series and Google Earth 
(GSD 0.3--0.8\,m, image sizes from $1000 \times 1000$ to 
$10{,}000 \times 10{,}000$~pixels). It categorizes 
objects into 5 main classes and 37 sub-categories. Using 
only the ship-related data, our subset comprises 6,413 training images (37,997 OBB) and 2,932 validation images (27,703 OBB), offering diverse ship sub-types.

\subsection{ShipRSImageNet}
ShipRSImageNet~\cite{ShipRSImageNet} is a challenging 
dataset derived from multi-source optical satellites 
(WorldView-3, GF-2, JL-1), with a wide span in spatial 
resolution (0.12--6.0\,m) and diverse environmental 
conditions. We utilized its Level-2 hierarchy of 24 ship 
classes (excluding docks) to evaluate fine-grained 
classification, yielding 2,709 training images (11,834 OBB) and 692 validation images (3,459 OBB). The wide range of fine-grained 
categories provides a demanding benchmark for 
class-discriminative ship detection.

\begin{table*}[htbp]
\centering
\caption{Statistical analysis of ship major and minor axes across 
datasets. The central 95\% range reports the 2.5th and 97.5th 
percentile boundaries. The feature dilution rate 
$\delta_{minor}$~(\%) quantifies the fraction of a grid cell 
occupied by background along the minor axis for the smallest ships 
in each dataset, computed via Eq.~(\ref{eq:dilution_rate}) using 
the lower bound of the minor axis central 95\% range.}
\label{tab:axis_stats}
\resizebox{\textwidth}{!}{%
\begin{tabular}{@{}l cccc cccc cccc@{}}
\toprule
\multirow{2}{*}{\textbf{Dataset}} 
  & \multicolumn{4}{c}{\textbf{Major Axis (px)}} 
  & \multicolumn{4}{c}{\textbf{Minor Axis (px)}} 
  & \multicolumn{4}{c}{$\boldsymbol{\delta_{minor}}$ \textbf{(\%)}} \\
  \cmidrule(lr){2-5} \cmidrule(lr){6-9} \cmidrule(l){10-13}
  & \textbf{Min} & \textbf{Mean} & \textbf{Max} 
  & \textbf{Cent. 95\%}
  & \textbf{Min} & \textbf{Mean} & \textbf{Max} 
  & \textbf{Cent. 95\%}
  & \textbf{P5} & \textbf{P4} & \textbf{P3} & \textbf{P2} \\ 
\midrule
SODA-A 
  & 2.97 & 27.31 & 135.72 & $[8, 64]$
  & 1.88 & 10.07 & 37.58 & $[4, 32]$
  & 87.5 & 75.0 & 50.0 & 0.0 \\
DOTA-v1.5 
  & 5.20 & 48.93 & 1783.64 & $[8, 128]$
  & 1.40 & 16.67 & 365.99 & $[4, 64]$
  & 87.5 & 75.0 & 50.0 & 0.0 \\
FAIR1M 
  & 1.00 & 49.74 & 1022.38 & $[8, 256]$
  & 1.00 & 12.65 & 156.52 & $[4, 64]$
  & 87.5 & 75.0 & 50.0 & 0.0 \\
ShipRSImageNet 
  & 7.00 & 155.00 & 1039.00 & $[16, 1024]$
  & 3.00 & 30.00 & 504.00 & $[8, 256]$
  & 75.0 & 50.0 & 0.0 & 0.0 \\
\midrule
\textbf{Overall} 
  & \textbf{--} & \textbf{70.24} & \textbf{--} 
  & $\boldsymbol{[8, 256]}$
  & \textbf{--} & \textbf{17.34} & \textbf{--} 
  & $\boldsymbol{[4, 64]}$
  & \textbf{87.5} & \textbf{75.0} & \textbf{50.0} 
  & \textbf{0.0} \\
\bottomrule
\end{tabular}%
}
\end{table*}

\begin{figure*}[!ht]
    \centering
    \includegraphics[width=\textwidth]{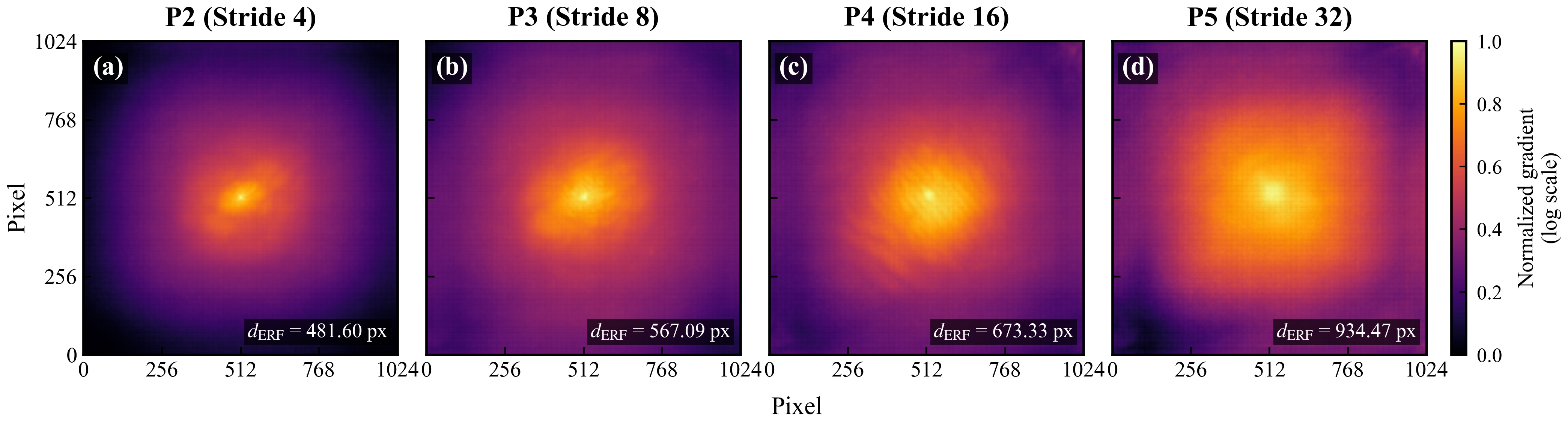}
    \caption{ERF of YOLOv9-E measured at four pyramid levels following the methodology of Luo et al.~\cite{ERF}, averaged over 100 random Gaussian inputs. (a)~P2 (stride~4), (b)~P3 (stride~8), (c)~P4 (stride~16), (d)~P5 (stride~32). Color intensity represents log-scaled normalized gradient magnitude. The ERF diameter $d_{\mathrm{ERF}}$, defined as the equivalent circular diameter enclosing 99.7\% of cumulative gradient energy ($3\sigma$ level), is 481.60, 567.09, 673.33, and 934.47~pixels for P2 through P5, respectively. Even at P4, the ERF already exceeds 2.6 times the 97.5th percentile of the observed ship major axis (256~px, Table~\ref{tab:axis_stats}), indicating that P5 provides marginal additional coverage of ship-relevant spatial extent.}
\label{fig:ERF}
\end{figure*}

\section{Problem Definition}
\label{sec:problem_definition}

The conventional P3--P5 head presents two structural 
mismatches with the empirical scale distribution of ships 
in satellite imagery. To characterize this distribution, 
we conducted a morphometric analysis across the four 
datasets, deriving all statistics from the original OBB 
annotations to avoid distortions introduced by 
preprocessing or resizing. Unlike general objects in 
natural images, ships in satellite imagery exhibit 
extreme aspect ratios, so the major and minor axes are 
analyzed separately. Table~\ref{tab:axis_stats} 
summarizes the resulting statistics, and 
Fig.~\ref{fig:ship_size_distribution} visualizes the 
joint distributions as density heatmaps. We formalize the minor-axis mismatch as a feature dilution 
in Section~\ref{sec:prob_dilution}, and the major-axis 
mismatch as a receptive-field overshoot in 
Section~\ref{sec:prob_erf}. Together, these mismatches 
motivate the pyramid level shift proposed in 
Section~\ref{sec:proposed_method}.

\subsection{Feature Dilution along the Minor Axis}
\label{sec:prob_dilution}

The minor axis distribution is markedly narrower than the 
major axis distribution. The aggregate mean is merely 
17.34~pixels, and the central 95\% range spans the narrow 
interval $[4, 64]$~pixels. The lower bound of this range, 
the 2.5th percentile $L_{minor}^{2.5\%}$, is 4~pixels for 
SODA-A, DOTA-v1.5, and FAIR1M, and 8~pixels for 
ShipRSImageNet. Across these datasets, the 2.5th 
percentile falls below the stride of P4 and P5 in every 
case and below the P3 stride for three of the four 
datasets, producing a fundamental resolution deficit in 
the grid-based detection mechanism.

To quantify how much of a grid cell encodes background 
rather than the target at a given feature map stride, we 
introduce the \textit{feature dilution rate along the 
minor axis}, $\delta_{minor}$:
\begin{equation}
\delta_{minor} = \max\!\left(0,\; 1 - \frac{L_{minor}^{2.5\%}}{S}
\right) \times 100\%
\label{eq:dilution_rate}
\end{equation}
where $L_{minor}^{2.5\%}$ denotes the 2.5th percentile of 
the minor axis distribution and $S$ is the stride of the 
corresponding pyramid level. When $\delta_{minor} > 0$, 
the target's minor axis is narrower than one grid cell, 
and the fraction $\delta_{minor}$ of that cell encodes 
background rather than object features. The grid-based 
detection mechanism of YOLO assigns one prediction per 
feature-map cell, so this dilution directly degrades the 
spatial resolution at which the target can be 
represented.

Substituting these per-dataset percentiles into 
Eq.~(\ref{eq:dilution_rate}) yields the $\delta_{minor}$ 
values shown in the rightmost columns of 
Table~\ref{tab:axis_stats}. $\delta_{minor}$ reaches 
87.5\% at P5, 75.0\% at P4, and 50.0\% at P3 for SODA-A, 
DOTA-v1.5, and FAIR1M, with 0\% reached only at P2 
(stride 4). ShipRSImageNet, owing to its larger 2.5th 
percentile, reaches $\delta_{minor} = 0$ already at P3 
(stride 8). The conventional P3--P5 head configuration 
therefore systematically fails to resolve the narrowest 
ships in the distribution.

The condition $\delta_{minor} = 0$ should be interpreted 
as a necessary but not sufficient condition for accurate 
boundary regression. The Nyquist--Shannon sampling 
theorem~\cite{Shannon} requires a sampling interval 
strictly less than half of the smallest spatial feature 
to be reconstructed, which in the present context 
corresponds to $S < L_{minor} / 2$. For 
$L_{minor}^{2.5\%} = 4$ pixels, this would demand $S < 2$, 
a constraint that the P2 head ($S = 4$) does not satisfy. 
The P2 head instead satisfies the weaker condition 
$S \leq L_{minor}$, which guarantees that the target 
occupies at least one complete grid cell and therefore 
admits a valid prediction. We refer to this weaker 
condition as the spatial representability condition. 
Going below P2 would approach the strict Nyquist limit 
at the cost of dramatically increased computation, and we 
therefore adopt P2 as the operational lower bound and 
rely on the OBB regression head to recover sub-cell 
boundary detail.

These results establish P2 as the minimum pyramid level 
required to satisfy the spatial representability 
condition for the central 95\% of the observed ship 
scale distribution.

\subsection{Receptive Field Overshoot along the Major Axis}
\label{sec:prob_erf}

The major axis statistics are concentrated at small to 
moderate scales across all four datasets. As reported in 
the Overall row of Table~\ref{tab:axis_stats}, the 
aggregate mean is 70.24~pixels, and the upper bound of 
the central 95\% range is 256~pixels, meaning that 
approximately 97.5\% of all ship instances have a major 
axis shorter than 256~pixels. Per-dataset upper bounds 
range from 64~pixels (SODA-A) to 1024~pixels 
(ShipRSImageNet), reflecting the inclusion of large 
warships in the latter. However, even in this most 
heterogeneous dataset, the central 95\% upper bound 
remains below the maximum observed value 
(1039~pixels).

These patterns are visually corroborated by the joint 
distributions in Fig.~\ref{fig:ship_size_distribution}. 
Across all four datasets, ship instances cluster in the 
lower-left region of the heatmap. SODA-A, DOTA-v1.5, and 
FAIR1M show particularly tight clustering below 64\,px 
on the minor axis and 256\,px on the major axis.

A sufficiently large receptive field is necessary for 
capturing contextual information, but an excessively 
large one introduces background clutter and incurs 
unnecessary computation. We therefore examine whether 
the deepest pyramid level, P5, provides any context that 
is actually used in the detection of ship-sized targets.

The spatial coverage of a pyramid level is commonly 
estimated by its Theoretical Receptive Field (TRF), the 
analytical extent computed from kernel sizes and 
strides~\cite{araujo2019computing}. TRF, however, assumes that every pixel within the receptive 
field contributes equally to the output. Luo et al.~\cite{ERF} showed that, 
due to the multiplicative accumulation of weights along 
convolution paths, pixel contributions in deep networks 
follow an approximately Gaussian distribution. Central 
pixels participate in exponentially more paths than 
peripheral ones, and the ERF, the region that 
meaningfully influences the output, therefore occupies 
only a fraction of the TRF. Formally, the contribution 
of an input pixel at position $(i, j)$ to a feature 
vector $h_{u,v}$ can be quantified as
\begin{equation}
\text{ERF}(i, j) \propto \sum_{c} \left|
\frac{\partial h_{u,v,c}}{\partial x_{i,j}} \right|
\label{eq:erf}
\end{equation}
where $h_{u,v,c}$ denotes the activation of the $c$-th 
channel at position $(u, v)$ and $x_{i,j}$ is the input 
pixel intensity. We estimate the ERF empirically by 
averaging the absolute gradient of 
Eq.~(\ref{eq:erf}) over 100 random Gaussian inputs, 
following the protocol of Luo et al.

Fig.~\ref{fig:ERF} reports the measured ERF at each pyramid level of 
YOLOv9-E. The P4 head reaches an ERF diameter of 
approximately 673 pixels, already 2.6 times the 97.5th 
percentile of the observed ship major axis (256 pixels). 
The P5 head extends this to approximately 934 pixels, a 
further 38.8\% enlargement.

This 261-pixel diameter increase from P4 to P5 corresponds to the annular band lying between the P4 and P5 ERF radii (from 336.67 to 467.24 pixels). Computed directly on the measured P5 gradient map, the disk inside the P4 ERF radius already accounts for $95.6\%$ of the P5 ERF's cumulative gradient energy, so the annular band traversed only by P5 carries the remaining $4.4\%$. The additional pixels gained by P5 thus contribute only marginally to the detection output. Moreover, since the P4 ERF already exceeds the 97.5th percentile of the observed ship scales, the extra area covered by P5 falls beyond any ship in the central 95\% of the distribution, capturing background rather than ship-relevant context. This is the receptive-field overshoot anticipated in Section~\ref{sec:problem_definition}.

Together with the resolution deficit established in 
Section~\ref{sec:prob_dilution}, this overshoot 
pinpoints both ends of the conventional P3--P5 
configuration as misaligned with the empirical ship 
scale distribution. P3 is too coarse to satisfy the 
spatial representability condition for the narrowest 
2.5\% of ships, and P5 extends beyond the central 95\% of the ship scale 
distribution. Pruning P5 and adding 
P2 thus yields a P2--P4 detection head whose receptive 
field is commensurate with the ship scale distribution 
at both ends, simultaneously eliminating unnecessary 
computation and reducing background contamination.

\begin{figure*}[!ht]
\centering
\includegraphics[width=\textwidth]{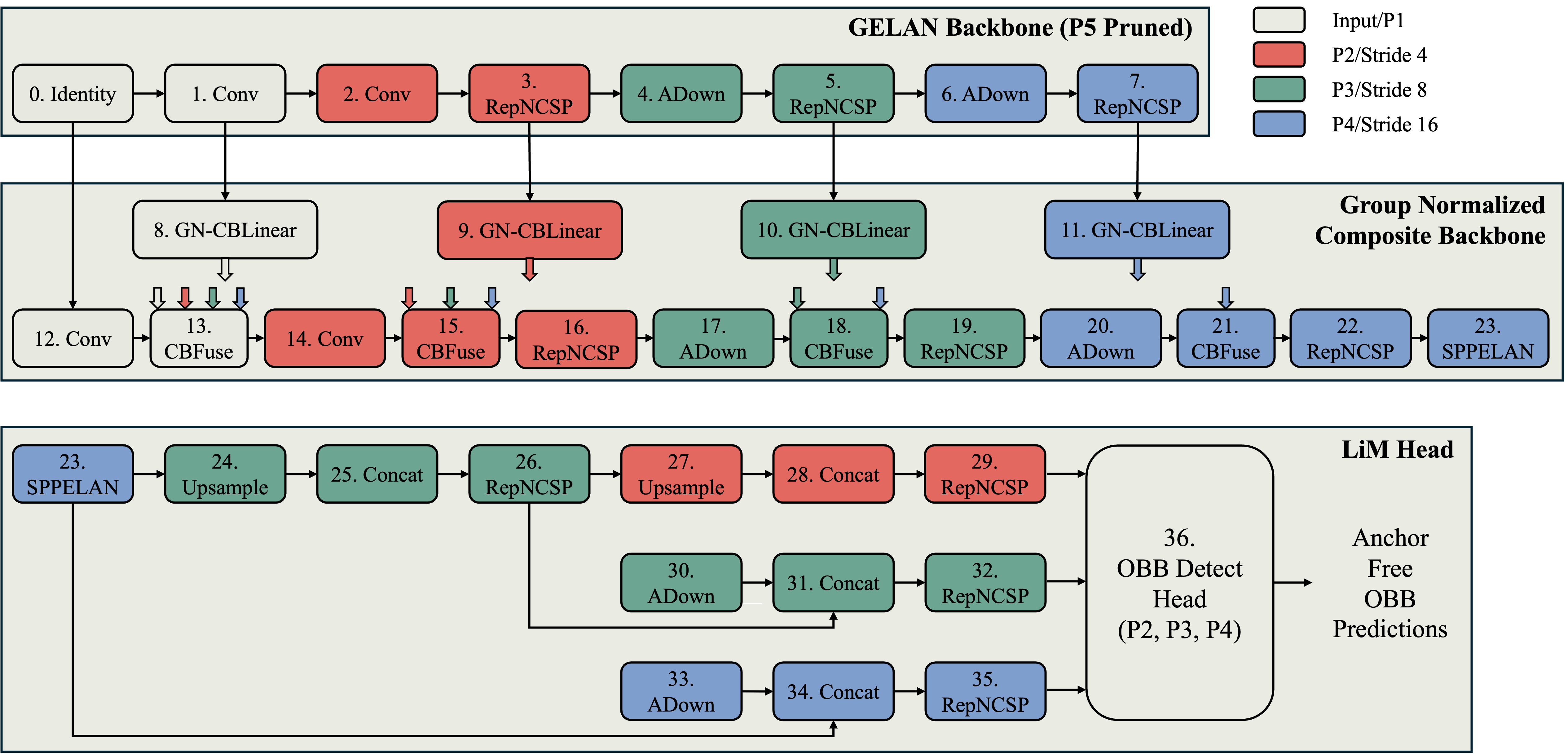}
\caption{\textbf{Overall architecture of the proposed LiM-YOLO.} 
The network consists of three parts. 
(Top)~The GELAN backbone extracts features from P1 through P4, 
with the P5 stage pruned to eliminate receptive field redundancy. 
(Middle)~The composite backbone projects multi-scale features through four GN-CBLinear modules (layers 8--11), each 
producing multi-level outputs that are fused via CBFuse operations. Colored arrows indicate the source 
pyramid level of each projection: white~(P1), red~(P2), 
green~(P3), and blue~(P4). The P4-level module (blue, layer~11) 
contributes to all four CBFuse operations, whereas the P1-level 
module (white, layer~8) contributes only to the first. 
(Bottom)~The LiM Head performs feature aggregation and 
multi-scale prediction at P2, P3, and P4 through an anchor-free 
OBB detection head.}
\label{fig:LiM-YOLO_architecture}
\end{figure*}

\begin{table*}[!h]
\centering
\caption{Details of the preprocessed datasets used in 
experiments. All annotations are in OBB format.}
\label{tab:dataset_composition}
\resizebox{\textwidth}{!}{%
\begin{tabular}{@{}lcccccc@{}}
\toprule
\textbf{Dataset} & \textbf{Source Platform} 
& \textbf{Resolution (GSD)} 
& \textbf{\# Train Img.} & \textbf{\# Train Inst.} 
& \textbf{\# Val Img.} & \textbf{\# Val Inst.} \\ 
\midrule
SODA-A & Google Earth & 0.5\,m -- 0.8\,m 
  & 1,030 & 37,971 & 323 & 21,908 \\
DOTA-v1.5 & Google Earth, GF-2, JL-1 & 0.3\,m -- 0.6\,m 
  & 2,657 & 56,313 & 572 & 11,474 \\
FAIR1M & Gaofen Series, Google Earth & 0.3\,m -- 0.8\,m 
  & 6,413 & 37,997 & 2,932 & 27,703 \\
ShipRSImageNet & WorldView-3, GF-2, JL-1 & 0.12\,m -- 6.0\,m 
  & 2,709 & 11,834 & 692 & 3,459 \\ 
\midrule
\textbf{Total} & \textbf{--} & \textbf{--} 
  & \textbf{12,809} & \textbf{144,115} 
  & \textbf{4,519} & \textbf{64,544} \\ 
\bottomrule
\end{tabular}%
}
\end{table*}

\section{Proposed Method}
\label{sec:proposed_method}

\subsection{Pyramid Level Shifted YOLOv9}

Conventional YOLO architectures adopt a pyramidal structure 
at P3, P4, and P5 (strides $2^3\!=\!8$, $2^4\!=\!16$, 
$2^5\!=\!32$) for multi-scale object detection. As 
established in Sections~\ref{sec:prob_dilution} and 
\ref{sec:prob_erf}, this configuration is misaligned with 
the empirical ship scale distribution at both ends. The P5 
level induces severe feature dilution along the minor axis, 
while its excessively large receptive field overshoots the 
observed major-axis scales.

Based on these findings, we propose a Pyramid Level Shift 
Strategy. First, we introduce a P2 head 
(stride~$2^2\!=\!4$) to capture high-resolution features. 
This configuration ensures $\delta_{minor} = 0$ for ships 
within the central 95\% of the observed scale distribution 
(Table~\ref{tab:axis_stats}), preserving the fine-grained 
spatial information of even the narrowest vessels. Second, 
we prune the P5 backbone and head, which were identified as 
structurally redundant in Section~\ref{sec:prob_erf}. 
Specifically, we remove the deepest backbone stage that 
downsamples the input by a factor of 32, together with the 
corresponding feature fusion blocks in the neck.

This redesign reduces the parameter count substantially 
while keeping the overall computational budget comparable, 
effectively trading the large-target capacity of P5 for a 
high-resolution P2 processing path. The resulting P2--P4 
architecture simultaneously achieves a lighter model and 
improved detection accuracy, as validated in
Section~\ref{sec:results}. The overall architecture of 
LiM-YOLO is illustrated in 
Fig.~\ref{fig:LiM-YOLO_architecture}.

\subsection{Group-Normalized CBLinear}

As data propagates through successive layers of a deep network,
the mutual information between the intermediate representations
and the target annotations is monotonically non-increasing.
This information bottleneck is captured by the Data Processing
Inequality:
\begin{equation}
\begin{split}
I(X, Y) &\ge I(f_1(X), Y) \ge I(f_2(f_1(X)), Y) \\
&\ge \dots \ge I(\hat{Y}, Y)
\end{split}
\label{eq:info_bottleneck}
\end{equation}
where $f_i$ denotes the transformation at the $i$-th layer,
$X$ is the input, $Y$ is the target, and $\hat{Y}$ is the
final prediction. Gradients backpropagated through many layers
therefore rely on progressively degraded information,
yielding suboptimal weight updates.

YOLOv9~\cite{YOLOv9} mitigates this through Programmable Gradient Information (PGI), which supplements the main branch during training with an auxiliary reversible branch and multi-level auxiliary information. Both are used only during training to improve the gradients that update the main branch, and neither is used at inference, so the inference network is the main branch alone and PGI adds no inference cost. Following~\cite{YOLOv9}, the reversible-branch idea 
is motivated by the property that, when a forward 
transformation $r(\cdot)$ admits an approximate inverse 
$v(\cdot)$ with $X \approx v(r(X))$, the mutual information 
is approximately retained:
\begin{equation}
X \approx v(r(X)) \;\Longrightarrow\; I(X, Y) \approx I(r(X), Y).
\label{eq:reversibility}
\end{equation}

In YOLOv9-E, the backbone additionally routes multi-scale features through the CBLinear and CBFuse modules, a CBNet-style composite backbone~\cite{liang2022cbnet} that operates within the backbone and is therefore retained at inference. The Ultralytics YOLOv9-E implementation on which we build retains this composite backbone but uses a single detection head, omitting the auxiliary reversible branch that the original YOLOv9 employs only during training~\cite{YOLOv9}. CBLinear applies a single
$1{\times}1$ convolution whose output channels equal the sum
of the target channel dimensions for each pyramid level, then
splits the result along the channel axis:
\begin{equation}
\text{CBLinear}(F) = \text{Split}\!\bigl(
\text{Conv}_{1\times1}(F),\; [c_1, c_2, \dots, c_K]\bigr)
\label{eq:cblinear_origin}
\end{equation}
where $[c_1, c_2, \dots, c_K]$ are the target channel
dimensions for $K$ pyramid levels. Each split tensor is
routed to the corresponding level of the composite backbone,
where CBFuse aggregates the contributions of multiple
backbone stages through nearest-neighbor resampling and
element-wise summation
(Fig.~\ref{fig:LiM-YOLO_architecture}). To minimize
distortion of the multi-level signal it propagates, the
original CBLinear contains neither normalization nor
non-linear activation. The $1{\times}1$ convolution is
applied with bias, yielding an affine projection.

However, this unnormalized design appears to be a source 
of training instability when applied to our task. Without 
normalization, the $1\times1$ projection has no mechanism 
to regulate the magnitude of its output activations, which 
therefore depend directly on the scale of the input feature 
maps and on the magnitude of the convolution weights. The composite backbone carries this unregulated signal into the main path through CBFuse. Consistent with this hypothesis, 
our ablation experiments show that introducing normalization into CBLinear yields consistent 
mAP improvements across all four datasets. Standard 
BN, however, is not a viable remedy when training under 
micro-batch constraints. Such constraints are common for 
high-resolution satellite imagery and force a batch size as 
small as two in our setup, well below the threshold at 
which BN statistics become unreliable~\cite{GN}.

We therefore propose the Group-Normalized CBLinear (GN-CBLinear) module. Following the standard CNN convention of applying normalization after 
convolution~\cite{GN}, we append GN immediately after the $1\times1$ 
convolution, before the channel-wise split. The convolution 
bias is omitted because the learnable shift parameter 
$\beta$ of GN serves the same role. This design remains 
compatible with the PGI philosophy because GN applies a per-channel affine transform that introduces less distortion. We continue to omit non-linear 
activation functions so that the gradient flow through the composite backbone remains less distorted.

GN divides the $C$ output channels of the convolution into 
$G$ groups of equal size and computes the mean $\mu_g$ and 
standard deviation $\sigma_g$ within each group 
independently of other samples in the batch. Because the 
statistics are derived from the spatial and intra-group 
channel dimensions of a single sample, GN remains stable 
regardless of batch size. The full operation of the 
proposed GN-CBLinear is defined as follows:
\begin{equation}
\begin{split}
Y &= \text{Conv}_{1\times1}(F) \\[2pt]
\hat{Y}_g &= \frac{Y_g - \mu_g}{\sqrt{\sigma_g^2 + \epsilon}}, \quad g = 1, \dots, G \\[2pt]
\tilde{Y} &= \gamma \odot [\hat{Y}_1, \dots, \hat{Y}_G] + \beta \\[4pt]
\text{GN-CBLinear}(F) &= \text{Split}\!\bigl(\tilde{Y},\; [c_1, \dots, c_K]\bigr)
\end{split}
\label{eq:gn_cblinear}
\end{equation}
where $Y_g$ denotes the $g$-th channel group of the 
convolution output, $\mu_g$ and $\sigma_g$ are its mean and standard deviation, $\gamma, \beta \in \mathbb{R}^{C}$ are per-channel learnable affine parameters, $\odot$ is channel-wise multiplication, $\epsilon$ is a small constant for 
numerical stability, and $[\hat{Y}_1, \dots, \hat{Y}_G]$ 
denotes concatenation along the channel axis to reconstruct 
the full normalized feature map. The outer Split operation 
distributes this feature map across pyramid levels, exactly 
as in the original CBLinear. In our implementation, we use 
$G\!=\!32$ groups.

This modification ensures that the composite backbone provides stable and consistent gradients throughout 
training, enabling effective optimization even under 
stringent memory constraints.

\section{Experimental Setup}

\subsection{Dataset Preparation and Preprocessing}

Using the four datasets introduced in 
Section~\ref{sec:dataset_analysis}, we applied a unified 
preprocessing pipeline to obtain the training and validation 
splits summarized in Table~\ref{tab:dataset_composition}.

We adopted a target patch size of $1024 \times 1024$ pixels. 
For images whose width or height exceeds this target, a 
sliding window strategy was applied to produce 
$1024 \times 1024$ patches. For the training set, an overlap 
of 256~pixels was used to reduce information loss from 
instance truncation at patch boundaries. Because of this overlap, instances near patch boundaries 
appear in multiple training patches, so the counts in 
Table~\ref{tab:dataset_composition} exceed the original 
(non-duplicated) counts shown in the 
Fig.~\ref{fig:ship_size_distribution} caption. The validation and 
test sets were cropped without overlap to preserve evaluation 
integrity. Images smaller than the target size were padded 
with zeros. Additionally, the training data were filtered to 
retain only those patches containing at least one ship 
instance.

From the multi-class datasets, only ship-related categories 
were retained: a single ``ship'' class from SODA-A and 
DOTA-v1.5, 9~ship sub-types from FAIR1M-v2.0, and the 
24~ship classes of the ShipRSImageNet Level-2 hierarchy 
(``Dock'' excluded).

\subsection{Implementation Details}

All experiments were conducted on a workstation equipped with 
a single NVIDIA RTX A6000 (48\,GB) GPU running Ubuntu~22.04. 
The implementation was built upon the Ultralytics YOLO 
framework~\cite{YOLOv8}. The batch size was set to 2, and 
the Adam optimizer was employed with an initial learning rate 
of $10^{-3}$, which was decayed to $10^{-5}$ over 100 epochs. The loss function configuration 
followed the default settings of YOLOv9~\cite{YOLOv9}.

To isolate the contributions of the proposed architectural 
modifications, the pyramid level shift strategy and 
GN-CBLinear, we established a stringent experimental protocol 
designed to eliminate confounding variables. First, we 
deliberately disabled the data augmentation techniques built 
into the Ultralytics framework. This ensures that any 
observed performance gains are attributable to structural 
changes rather than augmentation effects. Second, all models 
were trained from scratch with random initialization, without 
using weights pretrained on ImageNet or other external 
datasets. This protocol allows a fair assessment of each 
architecture's intrinsic feature extraction capability on 
remote sensing data.

All models, including the baseline, are built on this same Ultralytics implementation, so the ablations isolate the effect of each architectural change.

\subsection{Evaluation Metrics}

The performance evaluation was conducted from two 
complementary perspectives, detection accuracy and 
computational efficiency.

For detection accuracy, we adopted Mean Average Precision 
(mAP), a standard metric in object detection. We report both 
mAP\textsubscript{50} (at an IoU threshold of 0.5) and mAP\textsubscript{50:95} 
(averaged over IoU thresholds from 0.5 to 0.95 in steps of 
0.05). These metrics assess not only object localization but 
also the precision of orientation estimation in the OBB 
setting. We additionally report Precision and Recall to 
provide a more complete picture of detection behavior.

For computational efficiency, we measured Giga 
Floating-Point Operations (GFLOPs), the number of trainable 
parameters, and the pure inference time per image (in 
milliseconds), excluding pre- and post-processing overhead.

\begin{table*}[htbp]
\centering
\caption{Ablation study of head architecture and modules on the SODA-A dataset. The inference time is measured as the average time per image. The proposed LiM-YOLO is shown in bold. The best performance result in each column is in blue.}
\label{tab:ablation_soda}
\resizebox{0.95\textwidth}{!}{%
\begin{tabular}{ccccc|ccc|ccccc}
\toprule
\multicolumn{5}{c|}{\textbf{Components}} & \multicolumn{3}{c|}{\textbf{Efficiency}} & \multicolumn{5}{c}{\textbf{Performance (SODA-A)}} \\ \cmidrule(r){1-5} \cmidrule(lr){6-8} \cmidrule(l){9-13}
\thead{\textbf{P2} \\ \textbf{Head}} & \thead{\textbf{P3} \\ \textbf{Head}} & \thead{\textbf{P4} \\ \textbf{Head}} & \thead{\textbf{P5 Head} \\ \textbf{\& Backbone}} & \thead{\textbf{GN-} \\ \textbf{CBLinear}} & \thead{\textbf{Params} \\ \textbf{(M)}} & \thead{\textbf{GFLOPs}} & \thead{\textbf{Time} \\ \textbf{(ms/img)}} & \textbf{F1} & \textbf{Prec.} & \textbf{Rec.} & \textbf{mAP\textsubscript{50}} & \textbf{mAP\textsubscript{50:95}} \\ \midrule
 & \checkmark & \checkmark & \checkmark &  & 58.99 & 196.4 & 24.1 & 0.828 & 0.906 & 0.763 & 0.849 & 0.637 \\
\checkmark & \checkmark & \checkmark & \checkmark &  & 57.41 & 230.2 & 29.9 & 0.833 & \textcolor{best}{0.909} & 0.769 & 0.855 & 0.656 \\
\checkmark & \checkmark & \checkmark &  &  & 21.16 & 189.4 & 25.9 & \textcolor{best}{0.836} & 0.907 & \textcolor{best}{0.775} & 0.856 & 0.660 \\
\checkmark & \checkmark & \checkmark &  & \checkmark & \textbf{21.16} & \textbf{189.4} & \textbf{26.9} & \textbf{0.829} & \textbf{0.905} & \textbf{0.765} & \textcolor{best}{\textbf{0.861}} & \textcolor{best}{\textbf{0.662}} \\
\checkmark & \checkmark &  &  &  & 16.15 & 173.4 & 24.5 & 0.832 & 0.907 & 0.769 & 0.860 & 0.660 \\
\bottomrule
\end{tabular}%
}
\end{table*}

\begin{table*}[htbp]
\centering
\caption{Ablation study of head architecture and modules on the DOTA-v1.5 dataset.}
\label{tab:ablation_dota}
\resizebox{0.95\textwidth}{!}{%
\begin{tabular}{ccccc|ccc|ccccc}
\toprule
\multicolumn{5}{c|}{\textbf{Components}} & \multicolumn{3}{c|}{\textbf{Efficiency}} & \multicolumn{5}{c}{\textbf{Performance (DOTA-v1.5)}} \\ \cmidrule(r){1-5} \cmidrule(lr){6-8} \cmidrule(l){9-13}
\thead{\textbf{P2} \\ \textbf{Head}} & \thead{\textbf{P3} \\ \textbf{Head}} & \thead{\textbf{P4} \\ \textbf{Head}} & \thead{\textbf{P5 Head} \\ \textbf{\& Backbone}} & \thead{\textbf{GN-} \\ \textbf{CBLinear}} & \thead{\textbf{Params} \\ \textbf{(M)}} & \thead{\textbf{GFLOPs}} & \thead{\textbf{Time} \\ \textbf{(ms/img)}} & \textbf{F1} & \textbf{Prec.} & \textbf{Rec.} & \textbf{mAP\textsubscript{50}} & \textbf{mAP\textsubscript{50:95}} \\ \midrule
 & \checkmark & \checkmark & \checkmark &  & 58.99 & 196.4 & 24.6 & 0.883 & \textcolor{best}{0.942} & 0.831 & 0.913 & 0.736 \\
\checkmark & \checkmark & \checkmark & \checkmark &  & 57.41 & 230.2 & 29.9 & 0.883 & 0.936 & 0.836 & 0.915 & 0.738 \\
\checkmark & \checkmark & \checkmark &  &  & 21.16 & 189.4 & 25.8 & 0.891 & 0.940 & 0.847 & 0.923 & 0.744 \\
\checkmark & \checkmark & \checkmark &  & \checkmark & \textbf{21.16} & \textbf{189.4} & \textbf{27.2} & \textcolor{best}{\textbf{0.892}} & \textbf{0.933} & \textcolor{best}{\textbf{0.853}} & \textcolor{best}{\textbf{0.925}} & \textcolor{best}{\textbf{0.750}} \\
\checkmark & \checkmark &  &  &  & 16.15 & 173.4 & 24.8 & 0.889 & 0.936 & 0.846 & 0.921 & 0.740 \\
\bottomrule
\end{tabular}%
}
\end{table*}

\begin{table*}[htbp]
\centering
\caption{Ablation study of head architecture and modules on the FAIR1M dataset.}
\label{tab:ablation_fair1m}
\resizebox{0.95\textwidth}{!}{%
\begin{tabular}{ccccc|ccc|ccccc}
\toprule
\multicolumn{5}{c|}{\textbf{Components}} & \multicolumn{3}{c|}{\textbf{Efficiency}} & \multicolumn{5}{c}{\textbf{Performance (FAIR1M)}} \\ \cmidrule(r){1-5} \cmidrule(lr){6-8} \cmidrule(l){9-13}
\thead{\textbf{P2} \\ \textbf{Head}} & \thead{\textbf{P3} \\ \textbf{Head}} & \thead{\textbf{P4} \\ \textbf{Head}} & \thead{\textbf{P5 Head} \\ \textbf{\& Backbone}} & \thead{\textbf{GN-} \\ \textbf{CBLinear}} & \thead{\textbf{Params} \\ \textbf{(M)}} & \thead{\textbf{GFLOPs}} & \thead{\textbf{Time} \\ \textbf{(ms/img)}} & \textbf{F1} & \textbf{Prec.} & \textbf{Rec.} & \textbf{mAP\textsubscript{50}} & \textbf{mAP\textsubscript{50:95}} \\ \midrule
 & \checkmark & \checkmark & \checkmark &  & 59.00 & 196.4 & 24.4 & 0.422 & 0.388 & 0.463 & 0.395 & 0.285 \\
\checkmark & \checkmark & \checkmark & \checkmark &  & 57.41 & 230.3 & 30.8 & 0.421 & 0.381 & 0.471 & 0.392 & 0.284 \\
\checkmark & \checkmark & \checkmark &  &  & 21.16 & 189.5 & 25.8 & 0.437 & 0.404 & 0.477 & 0.402 & 0.290 \\
\checkmark & \checkmark & \checkmark &  & \checkmark & \textbf{21.16} & \textbf{189.5} & \textbf{26.7} & \textcolor{best}{\textbf{0.447}} & \textcolor{best}{\textbf{0.416}} & \textbf{0.482} & \textcolor{best}{\textbf{0.418}} & \textcolor{best}{\textbf{0.302}} \\
\checkmark & \checkmark &  &  &  & 16.15 & 173.5 & 24.9 & 0.441 & 0.406 & \textcolor{best}{0.483} & 0.414 & 0.301 \\
\bottomrule
\end{tabular}%
}
\end{table*}

\begin{table*}[!t]
\centering
\caption{Ablation study of head architecture and modules on the ShipRSImageNet dataset.}
\label{tab:ablation_shiprs}
\resizebox{0.95\textwidth}{!}{%
\begin{tabular}{ccccc|ccc|ccccc}
\toprule
\multicolumn{5}{c|}{\textbf{Components}} & \multicolumn{3}{c|}{\textbf{Efficiency}} & \multicolumn{5}{c}{\textbf{Performance (ShipRSImageNet)}} \\ \cmidrule(r){1-5} \cmidrule(lr){6-8} \cmidrule(l){9-13}
\thead{\textbf{P2} \\ \textbf{Head}} & \thead{\textbf{P3} \\ \textbf{Head}} & \thead{\textbf{P4} \\ \textbf{Head}} & \thead{\textbf{P5 Head} \\ \textbf{\& Backbone}} & \thead{\textbf{GN-} \\ \textbf{CBLinear}} & \thead{\textbf{Params} \\ \textbf{(M)}} & \thead{\textbf{GFLOPs}} & \thead{\textbf{Time} \\ \textbf{(ms/img)}} & \textbf{F1} & \textbf{Prec.} & \textbf{Rec.} & \textbf{mAP\textsubscript{50}} & \textbf{mAP\textsubscript{50:95}} \\ \midrule
 & \checkmark & \checkmark & \checkmark &  & 59.01 & 196.5 & 24.6 & 0.527 & 0.514 & 0.541 & 0.516 & 0.414 \\
\checkmark & \checkmark & \checkmark & \checkmark &  & 57.42 & 230.4 & 30.0 & 0.514 & 0.496 & 0.534 & 0.526 & 0.415 \\
\checkmark & \checkmark & \checkmark &  &  & 21.17 & 189.6 & 26.1 & 0.536 & 0.515 & 0.558 & 0.534 & 0.428 \\
\checkmark & \checkmark & \checkmark &  & \checkmark & \textbf{21.17} & \textbf{189.6} & \textbf{26.9} & \textcolor{best}{\textbf{0.574}} & \textcolor{best}{\textbf{0.548}} & \textcolor{best}{\textbf{0.601}} & \textcolor{best}{\textbf{0.578}} & \textcolor{best}{\textbf{0.448}} \\
\checkmark & \checkmark &  &  &  & 16.16 & 173.6 & 25.2 & 0.515 & 0.499 & 0.532 & 0.524 & 0.325 \\
\bottomrule
\end{tabular}%
}
\end{table*}

\section{Results}
\label{sec:results}

In this section, we present experimental results 
validating the effectiveness of LiM-YOLO. The evaluation 
is organized into four parts. First, we conduct ablation 
studies across four datasets to isolate the contributions 
of the pyramid level shift strategy and the GN-CBLinear 
module. Second, we benchmark our model against 
state-of-the-art detectors on the Integrated Ship 
Detection Dataset. Third, we analyze class-wise 
performance on the 24 fine-grained categories of 
ShipRSImageNet to examine scale-dependent behavior. 
Finally, we present qualitative detection examples.



\begin{table*}[htbp]
\centering
\caption{Comparison with state-of-the-art models on the Integrated Ship Detection Dataset. The proposed LiM-YOLO is shown in bold. The best performance result in each column is in blue.}
\label{tab:sota_comparison}
\resizebox{0.95\textwidth}{!}{%
\begin{tabular}{l|ccc|ccccc}
\toprule
\multirow{2}{*}{\textbf{Models}} & \multicolumn{3}{c|}{\textbf{Efficiency}} & \multicolumn{5}{c}{\textbf{Accuracy (Integrated Dataset)}} \\ \cmidrule(lr){2-4} \cmidrule(l){5-9}
 & \thead{\textbf{Params} \\ \textbf{(M)}} & \textbf{GFLOPs} & \thead{\textbf{Time} \\ \textbf{(ms/img)}} & \textbf{F1} & \textbf{Prec.} & \textbf{Rec.} & \textbf{mAP\textsubscript{50}} & \textbf{mAP\textsubscript{50:95}} \\ \midrule
YOLOv8x & 69.47 & 263.9 & 17.8 & 0.777 & 0.825 & 0.734 & 0.816 & 0.566 \\
YOLOv10x & 30.78 & 166.9 & 18.0 & 0.756 & 0.811 & 0.708 & 0.796 & 0.543 \\
YOLO11x & 58.78 & 203.8 & 18.6 & 0.764 & 0.822 & 0.713 & 0.805 & 0.554 \\
YOLOv12x & 61.02 & 208.1 & 36.7 & 0.721 & 0.793 & 0.662 & 0.748 & 0.494 \\
RT-DETR-X & 70.38 & 278.2 & 19.8 & 0.755 & 0.819 & 0.699 & 0.793 & 0.545 \\ \midrule
\textbf{Ours (LiM-YOLO)} & \textbf{21.16} & \textbf{189.4} & \textbf{26.7} & \textcolor{best}{\textbf{0.791}} & \textcolor{best}{\textbf{0.839}} & \textcolor{best}{\textbf{0.748}} & \textcolor{best}{\textbf{0.832}} & \textcolor{best}{\textbf{0.600}} \\ \bottomrule
\end{tabular}%
}
\end{table*}


\begin{table*}[t]
\caption{Class-wise mAP\textsubscript{50:95} on ShipRSImageNet across 
five ablation configurations. Classes are sorted by ascending 
average object area. The Baseline is the YOLOv9-E model with detection heads at P3--P5. P2--P5 adds the P2 head. P2--P4 further removes the P5 head and backbone. LiM-YOLO is P2--P4 with GN-CBLinear. P2--P3 additionally removes the P4 head. The proposed LiM-YOLO is shown in bold. The best result in each row is in blue.}
\label{tab:class_performance}
\centering
\setlength{\tabcolsep}{3.5pt} 
\begin{tabular}{l|c|ccccc}
\toprule
\textbf{Class} & \textbf{Area (px$^2$)} & \textbf{Baseline (P3--P5)} & \textbf{P2--P5} & \textbf{P2--P4} & \textbf{LiM-YOLO} & \textbf{P2--P3} \\
\midrule
\textbf{Average (All)} & - & 0.414 & 0.415 & 0.428 & \textcolor{best}{\textbf{0.448}} & 0.325 \\
\midrule
\multicolumn{7}{l}{\textit{Extra-Small Objects (Area $< 2000$)}} \\
Motorboat & 469 & 0.107 & 0.136 & 0.120 & \textcolor{best}{\textbf{0.138}} & 0.112 \\
Sailboat & 529 & 0.033 & 0.096 & 0.080 & \textcolor{best}{\textbf{0.162}} & 0.111 \\
\midrule
\multicolumn{7}{l}{\textit{Small Objects (Area $2000 \sim 10000$)}} \\
Fishing Vessel & 2350 & 0.152 & \textcolor{best}{0.209} & 0.151 & \textbf{0.194} & 0.155 \\
Hovercraft & 2391 & \textcolor{best}{0.585} & 0.504 & 0.425 & \textbf{0.497} & 0.531 \\
Yacht & 2558 & 0.499 & 0.512 & 0.560 & \textcolor{best}{\textbf{0.582}} & 0.459 \\
Tugboat & 2562 & 0.305 & 0.332 & \textcolor{best}{0.351} & \textbf{0.311} & 0.246 \\
Other Merchant & 3658 & 0.062 & 0.089 & \textcolor{best}{0.113} & \textbf{0.112} & 0.087 \\
Patrol & 4228 & \textcolor{best}{0.606} & 0.450 & 0.437 & \textbf{0.483} & 0.364 \\
Other Ship & 5398 & 0.297 & 0.295 & 0.266 & \textcolor{best}{\textbf{0.306}} & 0.273 \\
Barge & 6448 & \textcolor{best}{0.111} & 0.064 & 0.090 & \textbf{0.108} & 0.078 \\
Submarine & 7247 & 0.588 & 0.581 & \textcolor{best}{0.622} & \textbf{0.618} & 0.568 \\
Other Warship & 8173 & 0.297 & 0.303 & 0.320 & \textcolor{best}{\textbf{0.325}} & 0.296 \\
\midrule
\multicolumn{7}{l}{\textit{Medium Objects (Area $10000 \sim 30000$)}} \\
Ferry & 10054 & 0.161 & 0.090 & 0.127 & \textcolor{best}{\textbf{0.165}} & 0.113 \\
Frigate & 14408 & 0.633 & 0.654 & \textcolor{best}{0.723} & \textbf{0.719} & 0.593 \\
Cruiser & 15041 & 0.714 & 0.737 & \textcolor{best}{0.777} & \textbf{0.760} & 0.587 \\
Container Ship & 21136 & 0.413 & 0.391 & 0.440 & \textcolor{best}{\textbf{0.466}} & 0.255 \\
Cargo & 22157 & 0.546 & 0.486 & \textcolor{best}{0.571} & \textbf{0.551} & 0.366 \\
Destroyer & 22382 & 0.781 & 0.779 & \textcolor{best}{0.807} & \textbf{0.768} & 0.628 \\
Oil Tanker & 26248 & 0.326 & 0.282 & 0.415 & \textcolor{best}{\textbf{0.442}} & 0.309 \\
Auxiliary Ship & 28422 & 0.499 & 0.460 & 0.490 & \textcolor{best}{\textbf{0.559}} & 0.312 \\
Commander & 29649 & 0.481 & 0.519 & 0.377 & \textcolor{best}{\textbf{0.586}} & 0.298 \\
\midrule
\multicolumn{7}{l}{\textit{Large Objects (Area $> 30000$)}} \\
Landing & 31394 & 0.635 & 0.663 & \textcolor{best}{0.705} & \textbf{0.685} & 0.408 \\
RoRo & 33481 & 0.475 & 0.616 & 0.645 & \textcolor{best}{\textbf{0.682}} & 0.484 \\
Aircraft Carrier & 112301 & 0.639 & \textcolor{best}{0.712} & 0.669 & \textbf{0.538} & 0.172 \\
\bottomrule
\end{tabular}
\end{table*}


\subsection{Ablation Studies on Head Architecture and Modules}

To verify the cross-source generalization of the proposed 
architecture, we performed a step-wise ablation study 
under consistent conditions across SODA-A, DOTA-v1.5, 
FAIR1M, and ShipRSImageNet. The quantitative results are summarized in Tables~\ref{tab:ablation_soda}, \ref{tab:ablation_dota}, \ref{tab:ablation_fair1m}, and \ref{tab:ablation_shiprs} for SODA-A, DOTA-v1.5, FAIR1M, and ShipRSImageNet, respectively.

We first evaluated the effect of naively appending a P2 
head to the baseline YOLOv9-E architecture (P3--P5). As 
shown in the second rows of the ablation tables, this 
``expansion-only'' approach yielded only marginal gains 
across all four datasets. For instance, the F1-score on 
SODA-A increased from 0.828 to just 0.833, and on 
DOTA-v1.5 it remained unchanged at 0.883. At the same 
time, GFLOPs rose from 196.4 to 230.2 and inference time 
increased by 21.5--26.2\%. These results confirm that simply 
appending pyramid levels without addressing the 
structural redundancy of the P5 level fails to achieve an 
effective accuracy-efficiency balance across multiple 
datasets.

When we applied the proposed pyramid level shift, 
introducing the P2 head while simultaneously pruning the 
P5 backbone and head, the model achieved substantial 
improvements in both performance and efficiency. As shown 
in the third rows, this configuration reduced the 
parameter count by 64.1\% (from 58.99\,M to 21.16\,M) 
while consistently outperforming the baseline across all 
four datasets. The two structural changes contributed 
unequally across datasets. Adding the P2 head yielded its 
largest mAP\textsubscript{50:95} gain on SODA-A, where 
mAP\textsubscript{50:95} rose from 0.637 to 0.656 (an 
absolute gain of 0.019). This is consistent with the SODA-A 
scale profile, the dataset with the smallest average ship 
size in our benchmark suite, where the higher spatial 
resolution of P2 most directly addresses the feature 
dilution analyzed in Section~\ref{sec:prob_dilution}. 
Pruning the P5 level then yielded its largest gain on 
ShipRSImageNet, where mAP\textsubscript{50:95} rose 
from 0.415 to 0.428 (an absolute gain of 0.013). This is 
consistent with the characteristics of ShipRSImageNet, 
the 24-class fine-grained benchmark in our suite, where 
the reduction in P5-induced background contamination 
plausibly contributes to better class discrimination 
(Section~\ref{sec:prob_erf}).

To determine the minimum viable pyramid configuration, we 
further removed the P4 head (P2--P3 only). As shown in the final rows, this led to substantial degradation in 
mAP\textsubscript{50:95} almost exclusively on ShipRSImageNet
(0.428 to 0.325). The scale-dependent 
nature of this degradation is analyzed in detail in 
Section~\ref{sec:scale_dependent_classification}, and 
validates our decision to retain P4.

Finally, the integration of the GN-CBLinear module 
provided consistent mAP gains across all four 
datasets. As shown in the fourth rows of Tables~\ref{tab:ablation_soda}--\ref{tab:ablation_shiprs}, 
mAP\textsubscript{50:95} improved from 0.428 to 0.448 on 
ShipRSImageNet, an absolute gain of 0.020 over the standard P2--P4 configuration. This confirms that introducing GN into the 
originally unnormalized CBLinear module stabilizes 
training under the micro-batch regime imposed by 
high-resolution inputs.


\begin{figure*}[!t]
    \centering
    \includegraphics[width=0.95\textwidth]{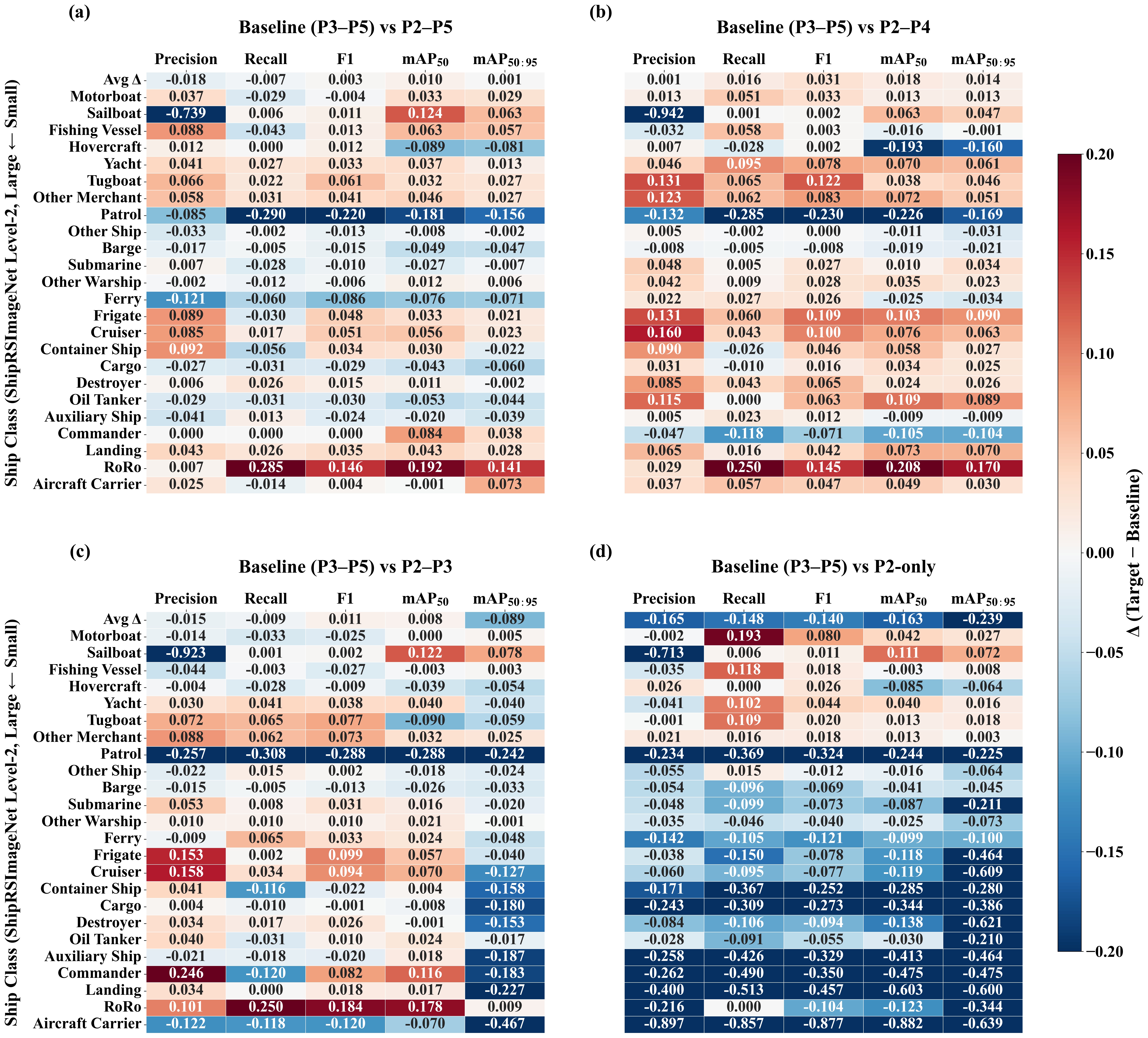}
    \caption{Performance difference ($\Delta$) heatmaps comparing 
    four ablation configurations against the Baseline (YOLOv9-E) 
    on ShipRSImageNet. The panels correspond to 
    (Top-Left)~P2--P5, which adds the P2 head. (Top-Right)~P2--P4, which further prunes the P5 backbone and head. (Bottom-Left)~P2--P3, which additionally removes the P4 head. (Bottom-Right)~the P2-only configuration, which keeps only the P2 head. The y-axis 
    lists 24 ship classes sorted by ascending average object area. 
    Red and blue cells indicate performance improvement 
    ($\Delta > 0$) and degradation ($\Delta < 0$), respectively.}
    \label{fig:combined_heatmaps}
\end{figure*}

\subsection{Comparison with State-of-the-Art Models}
\label{sec:sota}

To position LiM-YOLO within the current landscape of 
object detection, we evaluated it against leading 
detectors, including YOLOv8x, YOLOv10x, YOLO11x, 
YOLOv12x, and the transformer-based RT-DETR-X. All 
models were trained and evaluated on the Integrated Ship 
Detection Dataset, which merges the four benchmark 
datasets into a single corpus to provide a holistic 
assessment across diverse maritime conditions. The 
results are presented in 
Table~\ref{tab:sota_comparison}.

LiM-YOLO achieved the highest scores across all major accuracy metrics. It attained an mAP\textsubscript{50:95} of 0.600, surpassing the second-best model, YOLOv8x (0.566), by 0.034. Furthermore, it recorded the highest Precision (0.839) and Recall (0.748), indicating strong capability in detecting ships with few false positives and false negatives. Notably, LiM-YOLO surpasses YOLOv8x in mAP\textsubscript{50:95} 
by 0.034 while using only 30.5\% of its 
parameters (21.16\,M versus 69.47\,M). This performance gain 
is attributable to the pyramid level shift, which aligns 
the detection granularity with the physical scale of 
ships.

In terms of efficiency, LiM-YOLO requires only 21.16\,M 
parameters, approximately 30\% of RT-DETR-X (70.38\,M) 
and significantly fewer than YOLOv10x (30.78\,M). While 
its inference time (26.7\,ms) is slightly higher than 
some YOLO variants due to the processing of 
high-resolution P2 feature maps, it remains considerably 
faster than YOLOv12x (36.7\,ms). LiM-YOLO thus achieves 
state-of-the-art accuracy with significantly fewer 
parameters, validating that a well-targeted pyramid level 
shift can realize a ``Less is More'' balance between accuracy and efficiency.


\begin{figure}[!t]
\centering
\includegraphics[width=\linewidth]{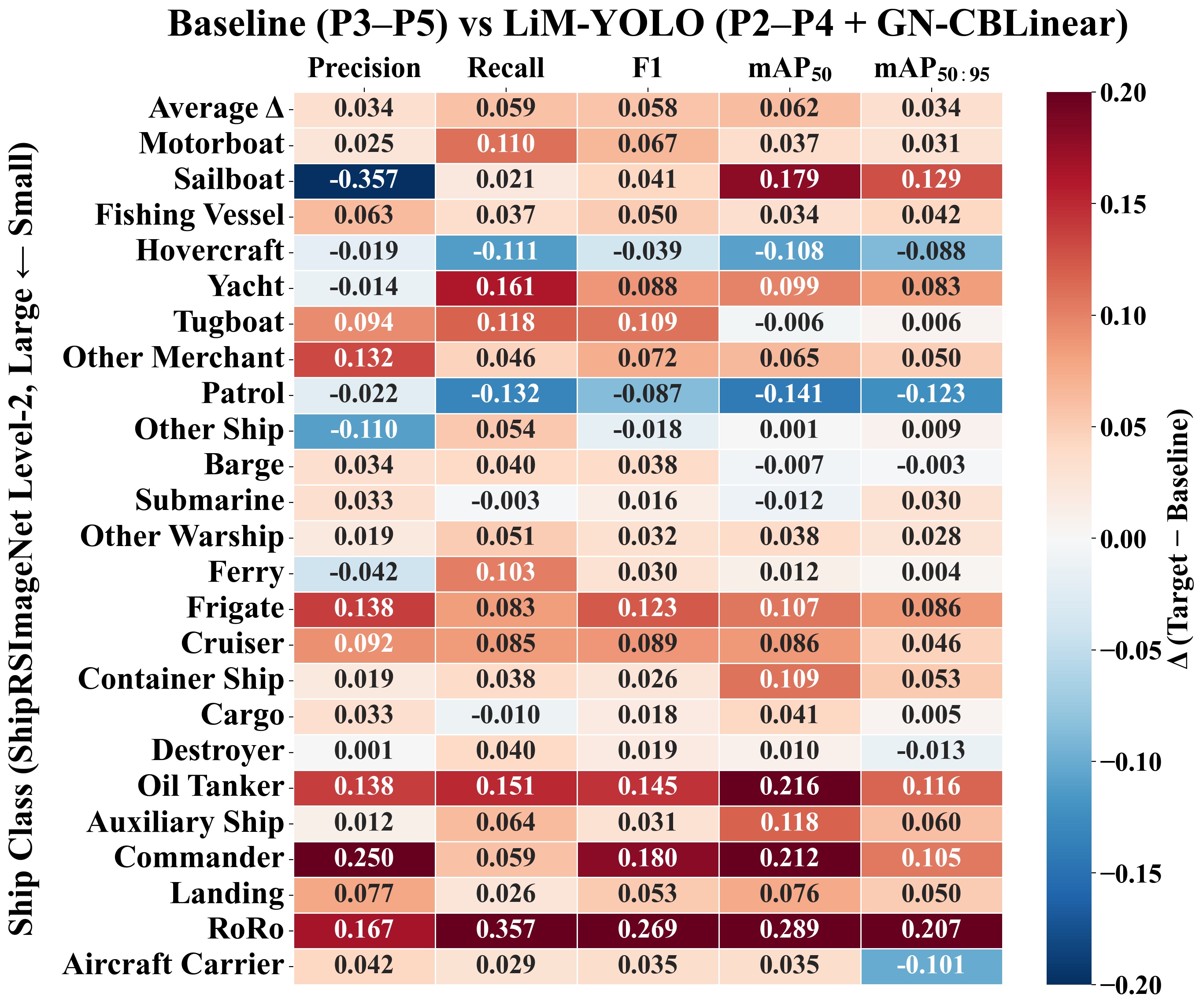}
\caption{\textbf{Class-wise performance difference ($\Delta$) 
between the Baseline and LiM-YOLO on ShipRSImageNet.} 
The heatmap displays differential values for Precision, Recall, 
F1-score, and mAP\textsubscript{50:95} across all ship categories sorted by 
size (small to large). Red cells ($\Delta > 0$) represent 
performance gains.}
\label{fig:heatmap_main}
\end{figure}


\begin{figure*}[!tp]
    \centering
    \setlength{\tabcolsep}{1.0pt}
    \renewcommand{\arraystretch}{0.3} 

    \begin{tabular}{c ccc}
        & \small \textbf{Baseline (YOLOv9-E)} & \small \textbf{LiM-YOLO (Ours)} & \small \textbf{Ground Truth} \\
        
        \rotatebox{90}{\small \textbf{SODA-A}} & 
        \includegraphics[width=0.29\linewidth]{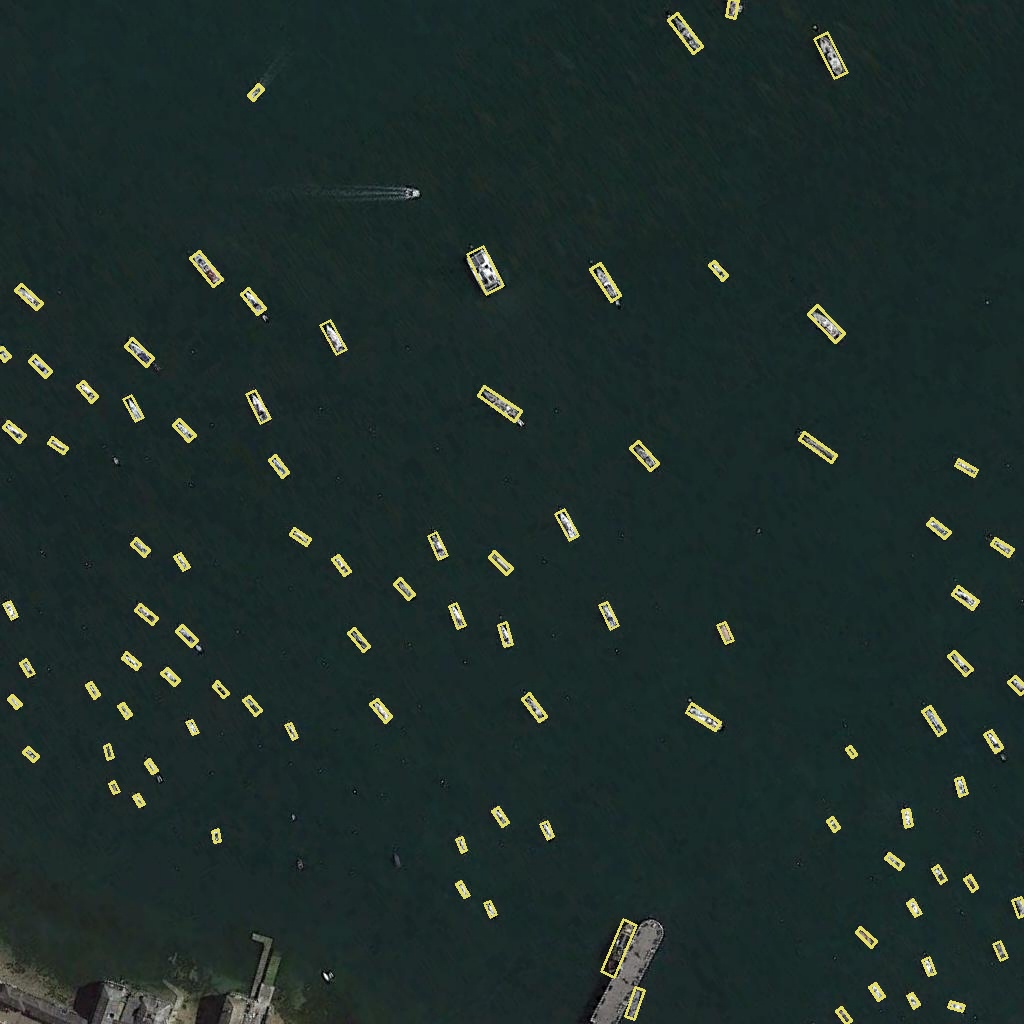} & 
        \includegraphics[width=0.29\linewidth]{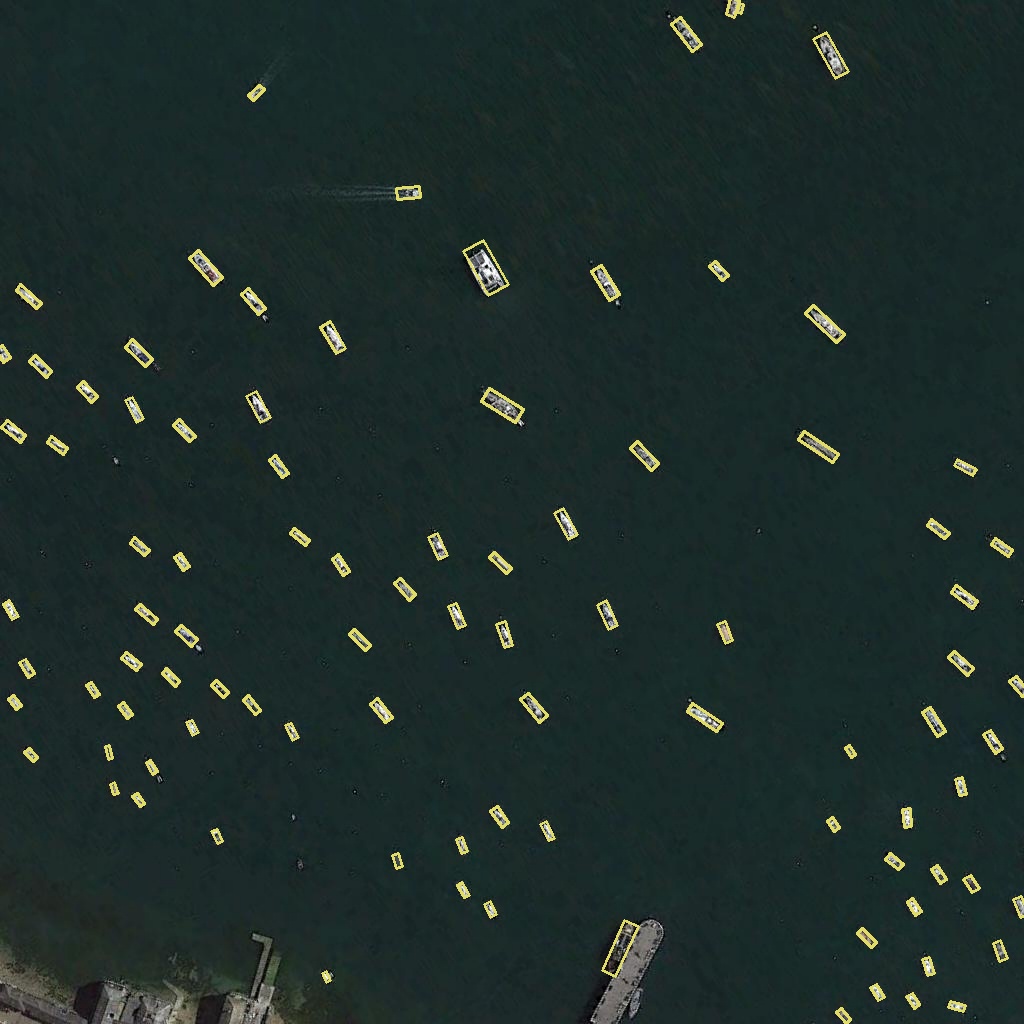} & 
        \includegraphics[width=0.29\linewidth]{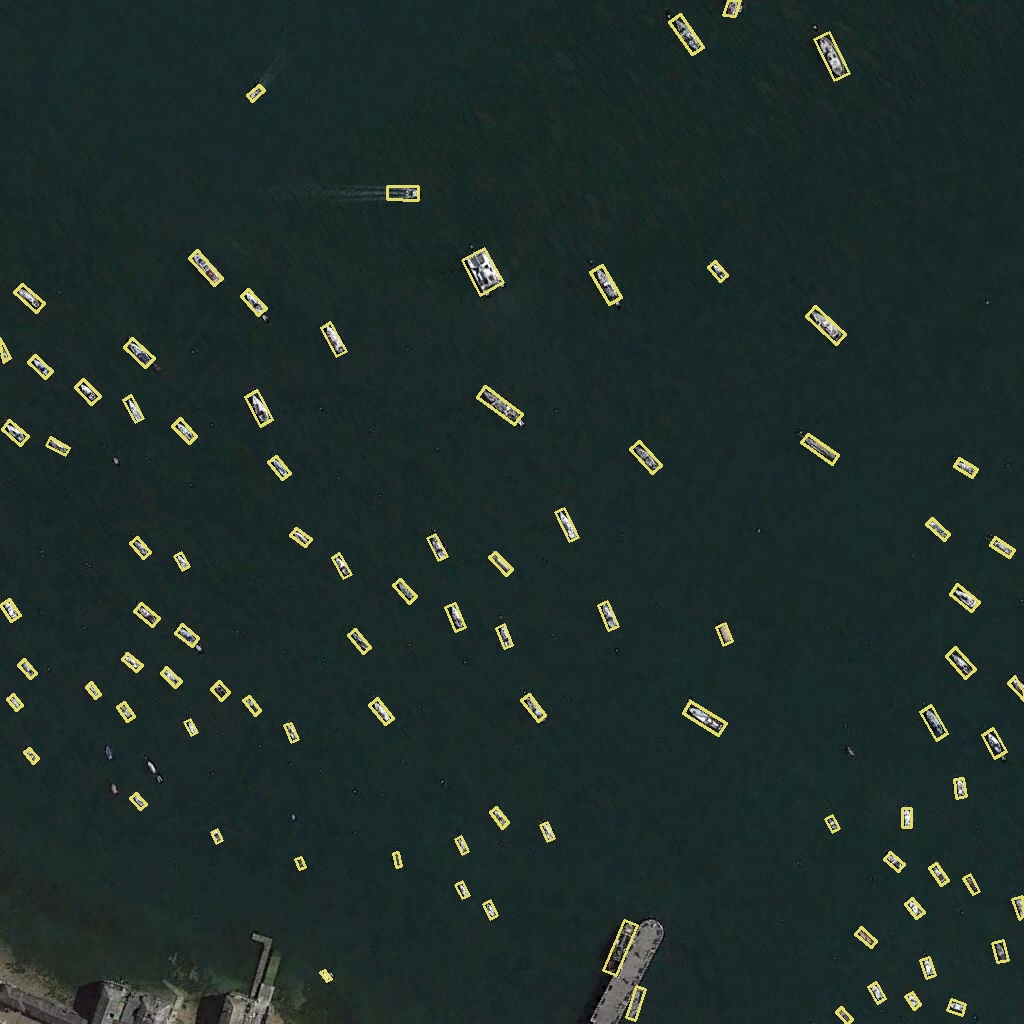} \\
        
        \rotatebox{90}{\small \textbf{DOTA-v1.5}} & 
        \includegraphics[width=0.29\linewidth]{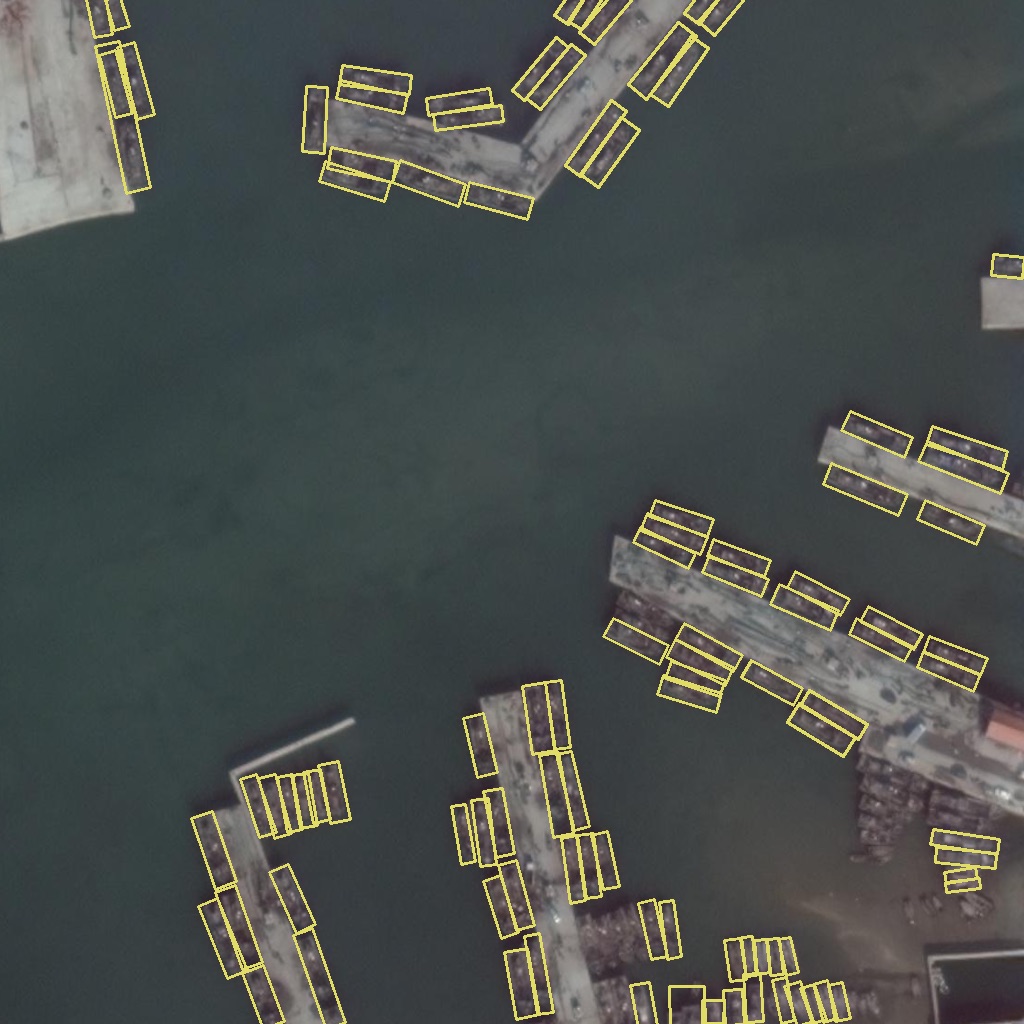} & 
        \includegraphics[width=0.29\linewidth]{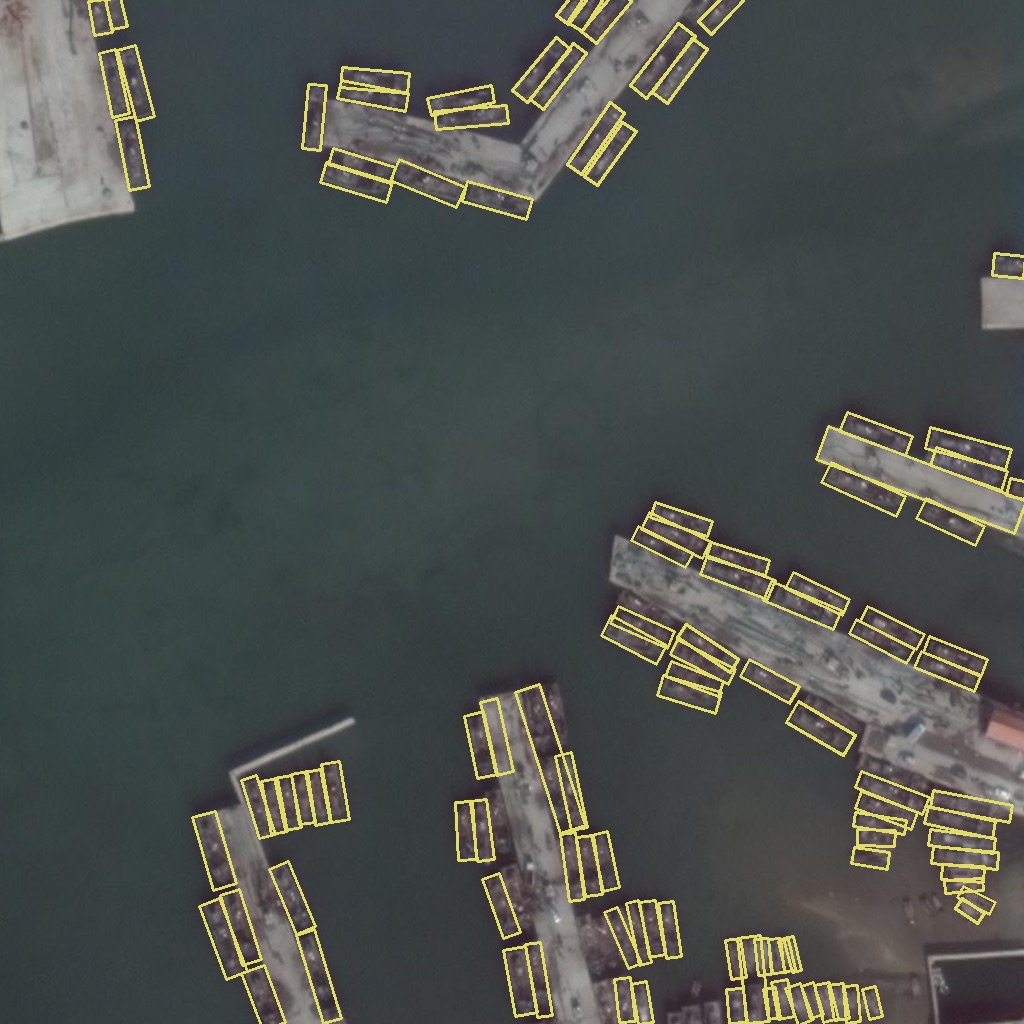} & 
        \includegraphics[width=0.29\linewidth]{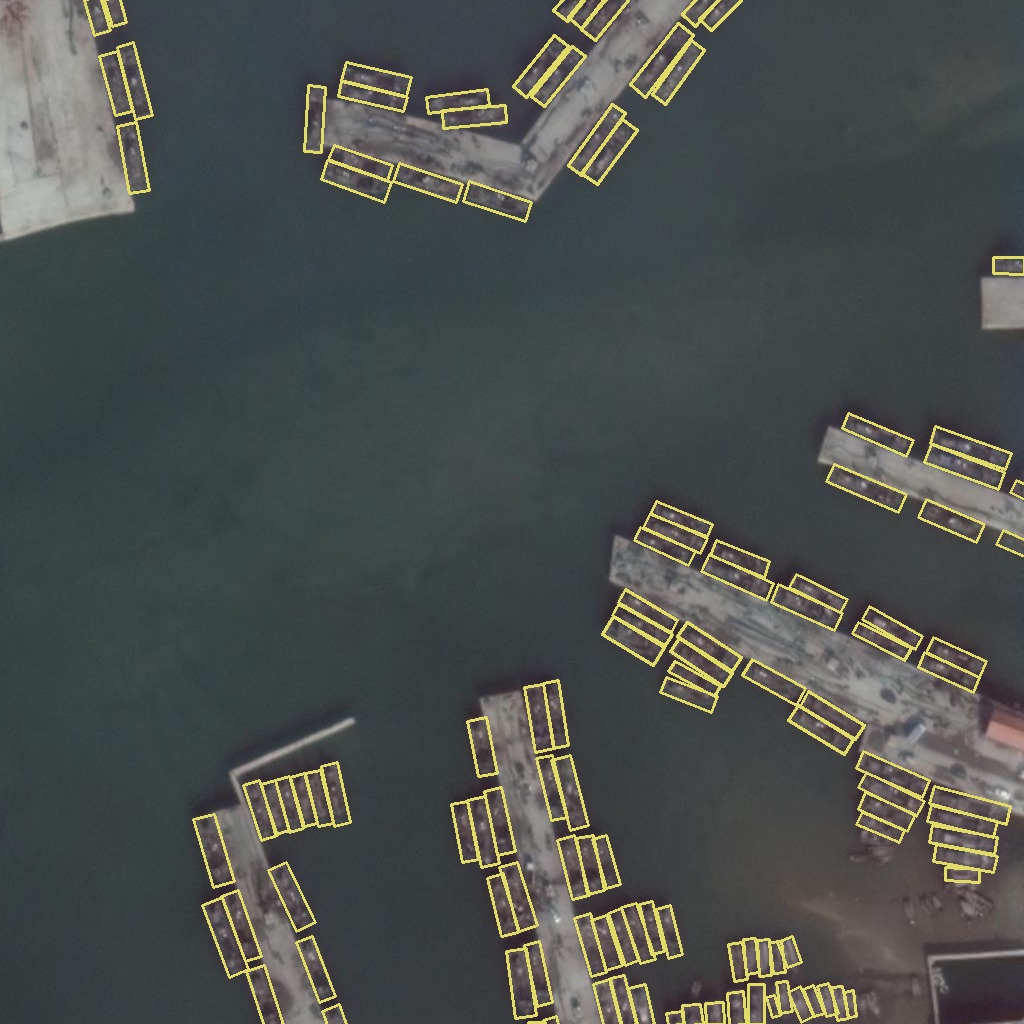} \\
        
        \rotatebox{90}{\small \textbf{FAIR1M}} & 
        \includegraphics[width=0.29\linewidth]{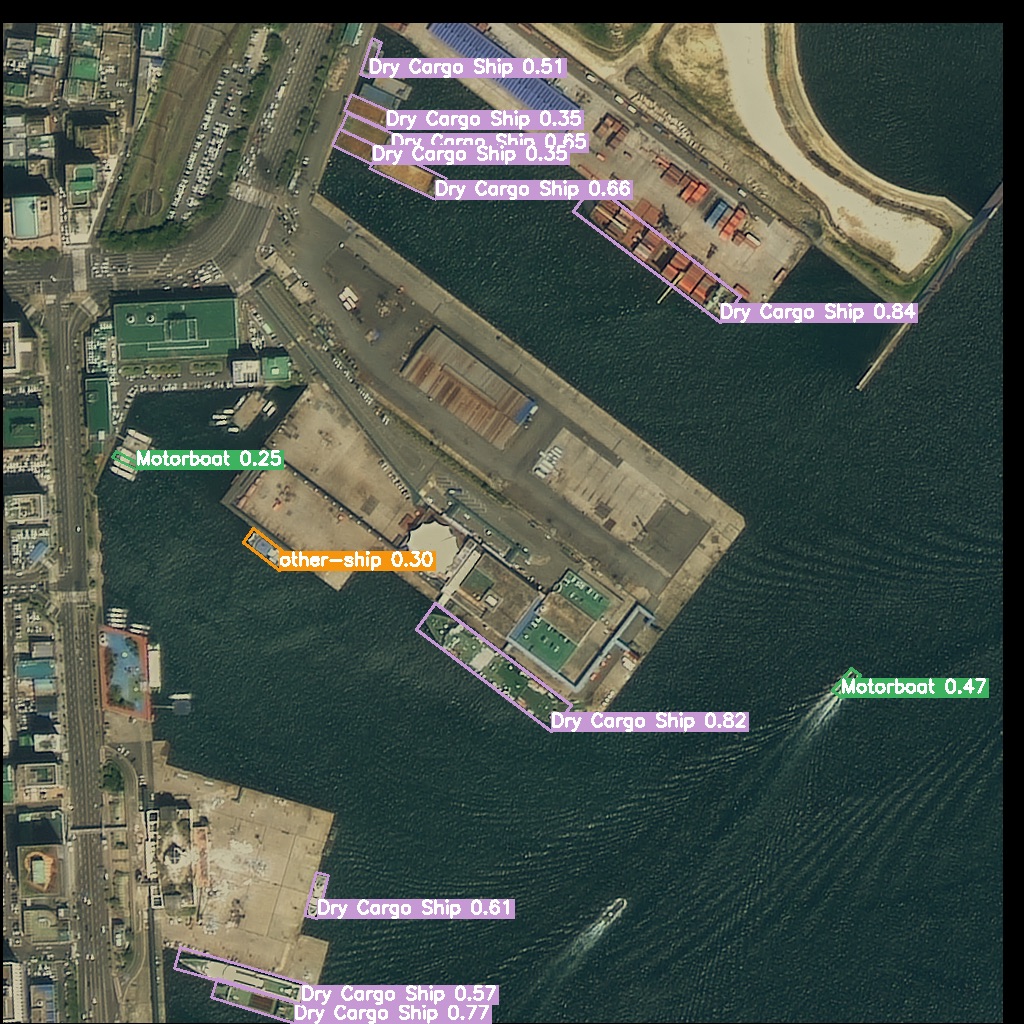} & 
        \includegraphics[width=0.29\linewidth]{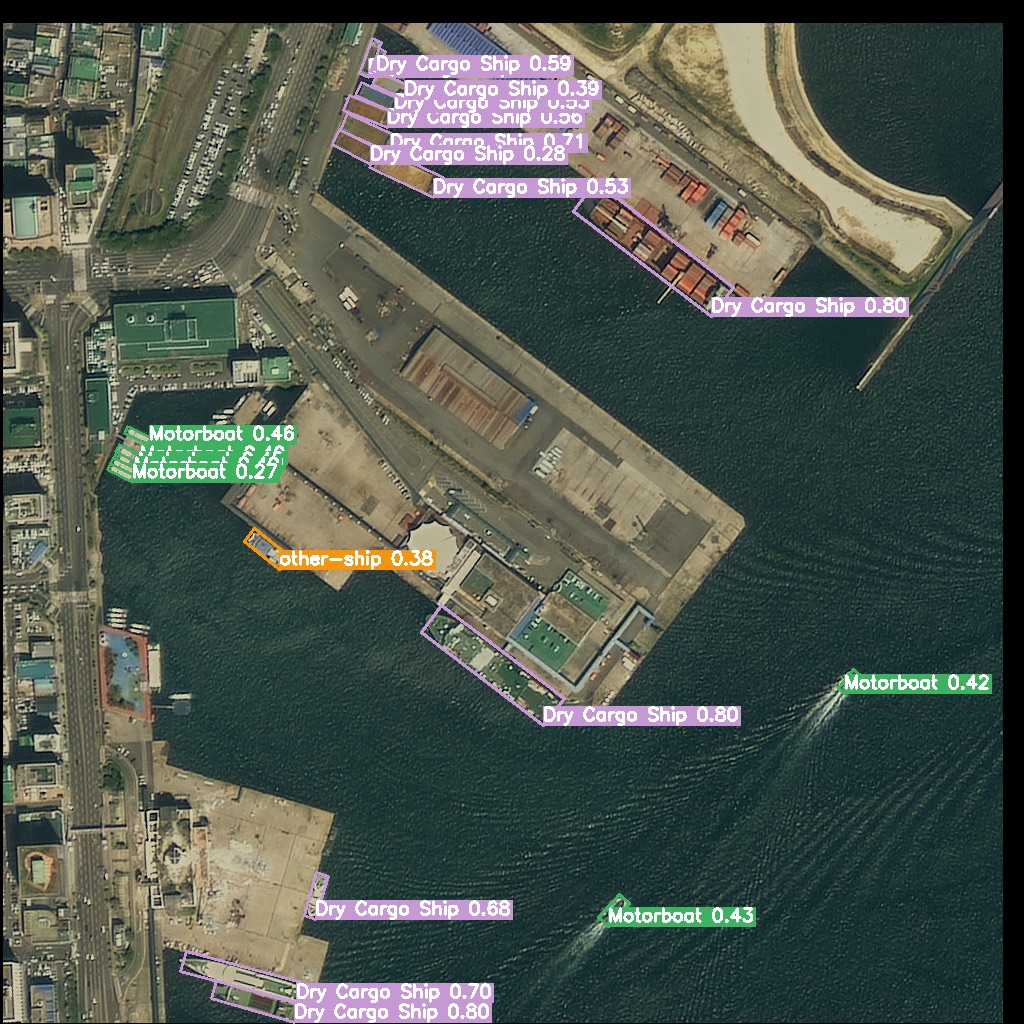} & 
        \includegraphics[width=0.29\linewidth]{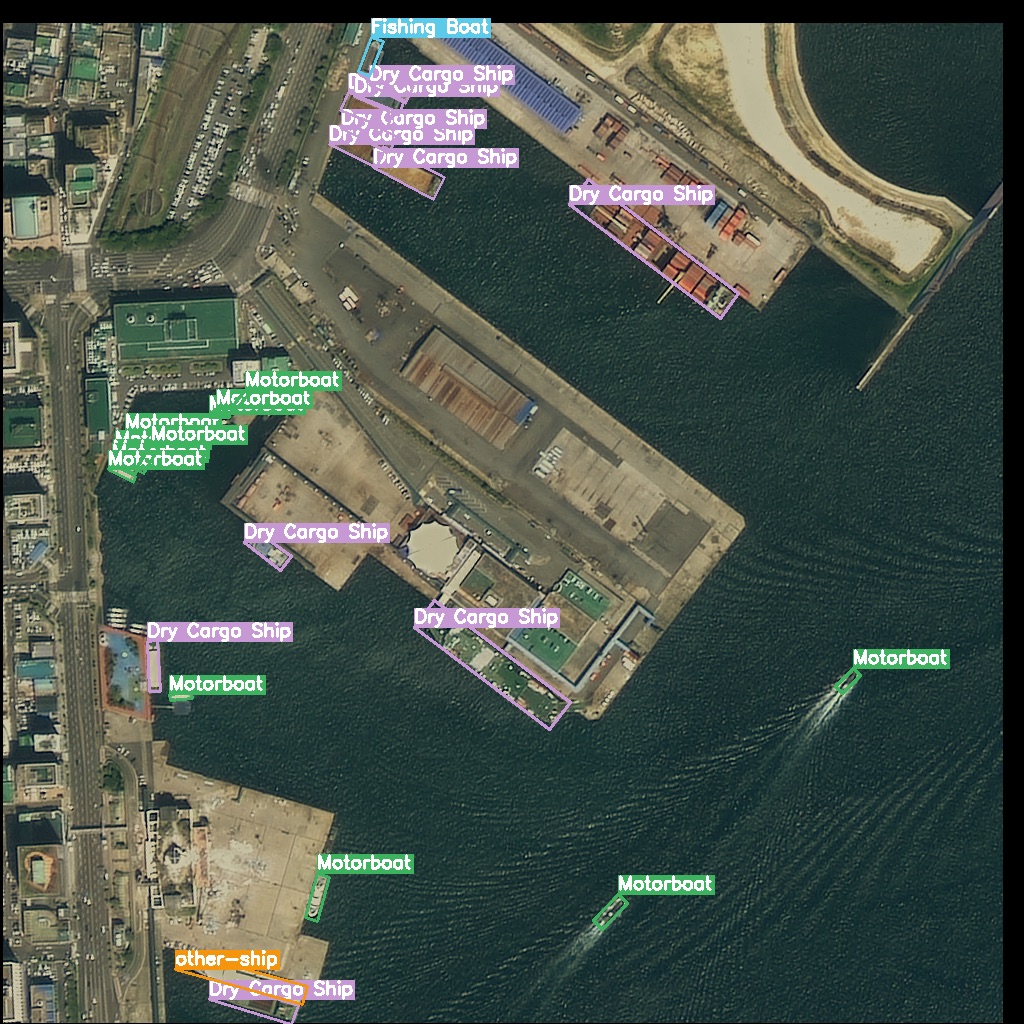} \\
        
        \rotatebox{90}{\small \textbf{ShipRSImageNet}} & 
        \includegraphics[width=0.29\linewidth]{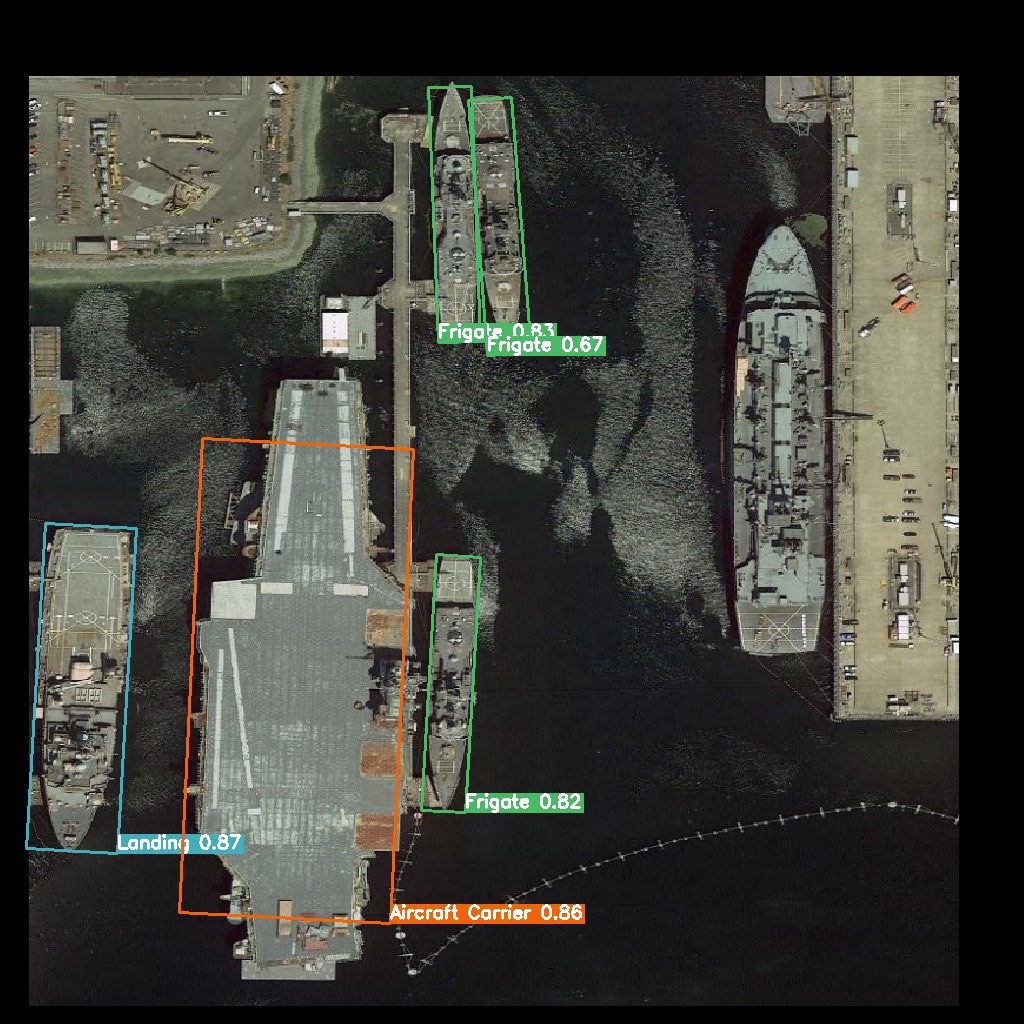} & 
        \includegraphics[width=0.29\linewidth]{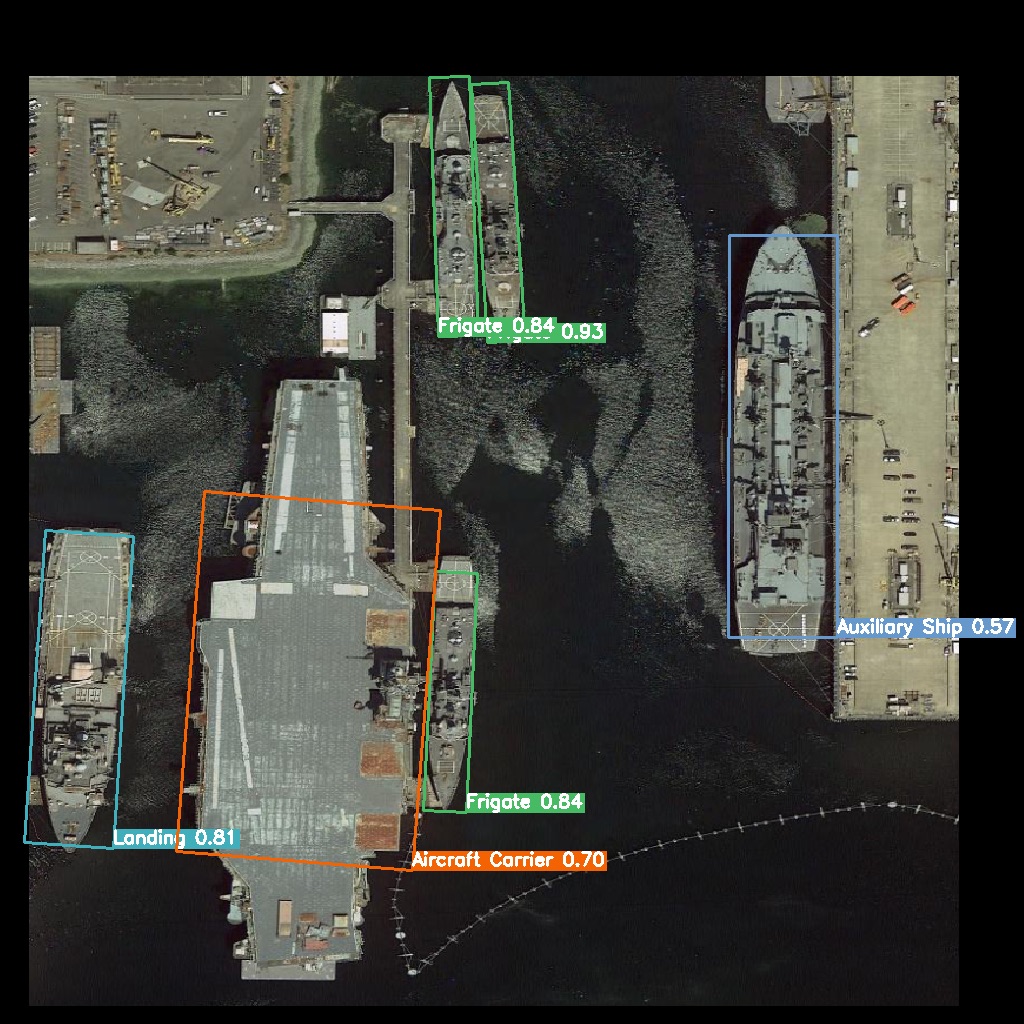} & 
        \includegraphics[width=0.29\linewidth]{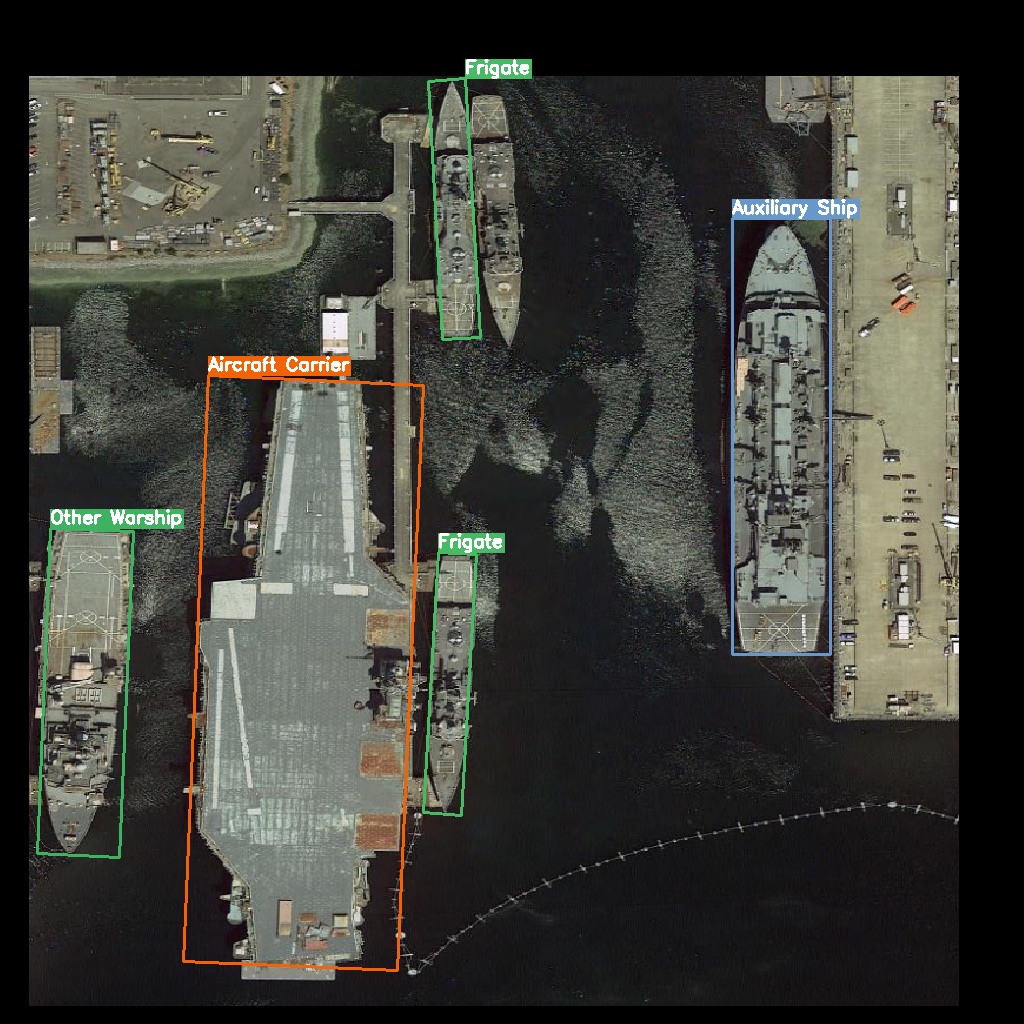} \\
    \end{tabular}
    
    \caption{\textbf{Qualitative comparison of detection results 
    across four remote sensing datasets.} Rows correspond to 
    SODA-A, DOTA-v1.5, FAIR1M-v2.0, and ShipRSImageNet (top to 
    bottom). Columns show the Baseline (YOLOv9-E), LiM-YOLO 
    (Ours), and Ground Truth. OBBs are overlaid on the images. For 
    datasets with a single ship class, class labels and confidence 
    scores are omitted for clarity.}
    \label{fig:qualitative_results}
\end{figure*}

\subsection{Scale-Dependent Classification Analysis}
\label{sec:scale_dependent_classification}

To examine how the core contributions of LiM-YOLO, namely the pyramid level shift and the GN-CBLinear module, affect 
detection performance across different object scales, we 
analyzed class-wise mAP\textsubscript{50:95} for all 24 ship categories 
in ShipRSImageNet, the most classification-challenging 
dataset in our study owing to its large number of 
fine-grained classes. 
Table~\ref{tab:class_performance} details the performance 
of each ablation configuration, with classes arranged by 
ascending average object area. The corresponding 
performance difference heatmaps are shown in 
Fig.~\ref{fig:combined_heatmaps}.

The impact of the P2 head (P2--P5) is most visible 
among the extra-small ship categories. The Baseline 
yielded mAP\textsubscript{50:95} of only 0.033 for 
Sailboat and 0.107 for Motorboat. P2--P5 improved 
these scores to 0.096 and 0.136, respectively. This 
substantial gain confirms that preserving high-resolution 
spatial details from the shallowest pyramid level is 
important for detecting the narrowest ships, 
whose features are lost during deep downsampling in the 
Baseline.

Adding P5 pruning to the P2 head (P2--P4) extends the 
benefit beyond the extra-small categories to medium and 
large ships. Relative to the Baseline, 
$\Delta$mAP\textsubscript{50:95} reaches $+0.17$ for RoRo, 
$+0.09$ for Frigate, $+0.09$ for Oil Tanker, and $+0.06$ 
for Cruiser, indicating that the reduction in 
receptive-field overshoot also helps medium-sized warships 
and merchant vessels. Although a few categories such as 
Patrol and Hovercraft show negative deltas, 16 of the 24 
classes exhibit positive $\Delta$mAP\textsubscript{50:95}, 
confirming that the benefit of P5 pruning is broadly 
distributed across the fine-grained class spectrum.

To determine the minimum viable pyramid depth, we 
removed the P4 head (P2--P3). The overall 
mAP\textsubscript{50:95} dropped from 0.428 to 0.325. 
As shown in Table~\ref{tab:class_performance}, this 
degradation is concentrated among the largest classes, 
with Aircraft Carrier plummeting from 0.669 to 0.172 and Landing dropping from 0.705 to 0.408. 
In contrast, extra-small classes such as Motorboat 
(0.120 $\to$ 0.112) were far less affected. This 
asymmetric pattern confirms that P4 is the minimum 
pyramid level required for semantic discrimination of 
large structures.

The full LiM-YOLO configuration, which augments P2--P4 
with the GN-CBLinear module, achieves the strongest 
class-wise behavior in Table~\ref{tab:class_performance}. 
Its overall average mAP\textsubscript{50:95} reaches 
0.448, an absolute improvement of 0.034 over the Baseline (0.414). It retains strong sensitivity 
to small targets, with Sailboat reaching 0.162, the 
highest among all configurations, while recovering 
accuracy for larger classes such as RoRo (0.682) and 
Commander (0.586) that were unstable in intermediate 
ablation models.

These improvements are visualized in 
Fig.~\ref{fig:heatmap_main}, which displays the per-class 
differential between LiM-YOLO and the Baseline. The 
figure is dominated by red cells ($\Delta > 0$), with the 
largest F1 gains observed in RoRo 
($\Delta$F1 $= +0.269$), Commander ($+0.180$), and 
Oil Tanker ($+0.145$). Among small-scale categories, 
Tugboat ($+0.109$) and Yacht ($+0.088$) also exhibit 
notable improvements, confirming the benefit of the 
high-resolution P2 level. The few classes showing 
degradation in mAP\textsubscript{50:95}, notably Aircraft 
Carrier ($-0.101$), degrade mainly at the GN-CBLinear addition. However, even for this class 
the F1-score improved slightly ($+0.035$), suggesting 
that the overall detection balance shifts favorably.

\subsection{Qualitative Results}
To provide intuitive validation of the proposed method, 
we visualized detection results on the test sets of the 
four datasets (Fig.~\ref{fig:qualitative_results}). For 
SODA-A and DOTA-v1.5, which contain a single ``ship'' 
class, class labels and confidence scores are omitted for 
clarity.

In the SODA-A result (first row), the Baseline failed to 
detect the small ship in the upper-center region. This failure 
is consistent with the feature dilution analyzed in 
Section~\ref{sec:prob_dilution}, where the high stride of 
P5 causes the features of tiny objects to be submerged in 
background content. LiM-YOLO, equipped with the P2 head 
that ensures $\delta_{minor} = 0$ for the vast majority of 
ships, successfully localized this target. In the 
DOTA-v1.5 result (second row), the Baseline struggled to 
resolve densely packed ships in the bottom-right area, 
whereas LiM-YOLO accurately distinguished individual 
instances, demonstrating that the P2--P4 configuration 
improves boundary discrimination in crowded scenes. In the FAIR1M result (third row), LiM-YOLO resolved a 
densely packed cluster of motorboats more clearly than 
the Baseline, and further detected a motorboat with a 
visible wake in the lower-central region of the scene 
that the Baseline missed.

In the ShipRSImageNet result (fourth row), the difference between the Baseline and LiM-YOLO becomes even more apparent. While the Baseline 
missed the medium-sized ``Auxiliary Ship,'' LiM-YOLO 
detected it correctly. This confirms that pruning P5 does 
not severely compromise the detection of larger objects, 
as the P4 ERF (approximately 673~pixels) provides 
sufficient contextual coverage for most ship categories. Most notably, LiM-YOLO 
detected a ``Frigate'' in the upper-center region with a 
higher confidence score than the Baseline, an instance 
that was omitted even from the ground truth annotations.

\section{Discussion}

Contemporary object detection research has generally 
assumed that deeper feature hierarchies yield better 
semantic abstraction. Our findings challenge this 
assumption in the domain of ship detection in optical 
remote sensing imagery. As shown in the ablation studies 
(Tables~\ref{tab:ablation_soda}--\ref{tab:ablation_shiprs}), 
the baseline YOLOv9-E, which retains the deep P5 level 
(stride $2^5\!=\!32$), consistently underperformed the 
proposed P2--P4 configuration across all four datasets. 
We interpret this as a consequence of architectural 
mismatch rather than insufficient model capacity. The 
``expansion-only'' strategies adopted by prior 
works~\cite{YOLO-RSA, YOLO-Ssboat}, which append 
additional pyramid levels while retaining P5, fail to 
address the root cause of feature dilution. Our ablation 
results make this point concrete. On SODA-A, naively 
appending a P2 head while keeping P5 (the expansion-only 
configuration) improved the F1-score from 0.828 to only 
0.833, yet once P5 was pruned, F1 rose to 0.836 with 
64\% fewer parameters. This pattern was consistent across 
all four datasets, demonstrating that pruning the P5 
level is as important as introducing the P2 level. 
Removing the deepest stage eliminates the background 
contamination introduced by excessive receptive fields 
and offsets much of the computational cost of processing 
the high-resolution P2 features, keeping the overall 
budget comparable to the baseline. This ``Less is More'' 
outcome shows that aligning the architecture with the 
target scale distribution is more effective than 
expansion-only strategies that simply add pyramid levels 
on top of the conventional P3--P5 head.

The qualitative success of LiM-YOLO in detecting densely 
packed small ships and narrow targets 
(Fig.~\ref{fig:qualitative_results}) directly supports the 
feature dilution analysis presented in 
Section~\ref{sec:prob_dilution}. By shifting the shallowest 
detection level from P3 to P2, LiM-YOLO satisfies the 
spatial representability condition $S \le L_{minor}$ for 
the central 95\% of the observed scale distribution, 
preserving the spatial cues required for accurate boundary 
regression. The detection of a ``Frigate'' missed even by 
the ground truth annotations in 
ShipRSImageNet~\cite{ShipRSImageNet} further illustrates 
the detection sensitivity of the P2--P4 configuration.

The integration of GN-CBLinear addressed a practical 
bottleneck that arises when training high-capacity models 
on high-resolution remote sensing imagery. Our ablation 
results confirm that incorporating GN into CBLinear yielded consistent mAP gains across all four 
datasets. This finding is relevant beyond ship detection. 
As sensor resolutions continue to improve and input sizes 
grow, the micro-batch constraint becomes increasingly 
prevalent, making batch-independent normalization 
increasingly important for training deep detectors on 
satellite data.

While LiM-YOLO achieves state-of-the-art ship detection 
accuracy with substantially fewer parameters, all 
experiments were conducted exclusively on ship detection 
datasets. Although the underlying principle of aligning 
the pyramid level configuration with the target scale 
distribution is general, its effectiveness on other 
remote sensing objects, such as vehicles or aircraft, 
remains to be empirically established. Extending the 
framework to a multi-category setting on aerial object 
detection benchmarks (e.g., DOTA across all categories) 
is an important direction for future work.

\section{Conclusion}

In this study, we identified a fundamental conflict between 
the small spatial extent of ships in optical remote sensing 
imagery and the conventional P3--P5 head configuration of 
standard YOLO architectures. Through a statistical analysis 
of ship scale distributions across four major remote 
sensing benchmarks, we showed that the conventional P5 
level induces severe feature dilution along the minor axis 
while its effective receptive field overshoots the 
major-axis scale of practically all observed ships.

To address these issues, we proposed LiM-YOLO, which 
shifts the feature pyramid from P3--P5 to P2--P4. The P2 
level (stride $2^2\!=\!4$) reduces $\delta_{minor}$ to 
zero across all four datasets, preserving the spatial 
integrity of even the narrowest ships. Simultaneously, 
pruning P5 eliminates receptive field redundancy and 
reduces overall computational overhead. We further introduced GN-CBLinear, which adds Group Normalization 
to the previously unnormalized linear projection in the composite backbone, stabilizing training under the micro-batch constraints 
imposed by high-resolution inputs.

Extensive experiments on SODA-A, DOTA-v1.5, FAIR1M-v2.0, 
and ShipRSImageNet confirmed the effectiveness of the 
proposed approach. On the Integrated Ship Detection 
Dataset, LiM-YOLO attained an mAP\textsubscript{50:95} 
of 0.600 with only 21.16\,M parameters, surpassing 
models up to three times its size. The class-wise 
analysis further demonstrated that LiM-YOLO substantially 
improves detection for the smallest ship categories, 
while maintaining competitive accuracy on larger targets.

These results establish that domain-specific 
architectural alignment, specifically a well-targeted 
pyramid level shift guided by target scale statistics, 
can outperform much larger detectors built on the 
conventional P3--P5 head configuration. LiM-YOLO 
embodies a ``Less is More'' principle for detector 
design. A detector matched to its target domain through 
a well-targeted pyramid level shift can be simultaneously 
more compact and more accurate than its conventional 
counterparts.

\section*{CRediT Authorship Contribution Statement}
\textbf{Seon-Hoon Kim:} Conceptualization, Data curation, Formal analysis, Investigation, Methodology, Resources, Software, Validation, Visualization, Writing -- original draft, Writing -- review \& editing.

\textbf{Yerin Kim:} Writing -- review \& editing.

\textbf{Hyeji Sim:} Data curation.

\textbf{Youeyun Jung:} Project administration.

\textbf{Okchul Jung:} Funding acquisition.

\textbf{Daewon Chung:} Supervision.

\section*{Funding}
This work was supported by the Korea Coast Guard through the Korea Institute of Marine Science \& Technology Promotion (KIMST) [grant number RS-2023-00238652, Integrated Satellite-based Applications Development for Korea Coast Guard].

\section*{Declaration of Competing Interest}
The authors declare that they have no known competing financial interests or personal relationships that could have appeared to influence the work reported in this paper.

\section*{Data Availability}
The source code supporting the findings of this study is publicly available at \url{https://github.com/egshkim/LiM-YOLO}. The benchmark datasets SODA-A, DOTA-v1.5, FAIR1M-v2.0, and ShipRSImageNet are publicly available from their respective original sources.

\ifCLASSOPTIONcaptionsoff
  \newpage
\fi



%
\bibliographystyle{IEEEtran}
\bibliography{references}

\end{document}